\documentclass[10pt,twocolumn,letterpaper]{article}

\usepackage{cvpr}              

\usepackage{graphicx}
\usepackage{amsmath}
\usepackage{amssymb}
\usepackage{booktabs}
\usepackage{microtype}

\usepackage[pagebackref,breaklinks,colorlinks]{hyperref}

\usepackage[capitalize]{cleveref}
\crefname{section}{Sec.}{Secs.}
\Crefname{section}{Section}{Sections}
\Crefname{table}{Table}{Tables}
\crefname{table}{Tab.}{Tabs.}

\usepackage{multirow}
\usepackage{amsmath,bm}
\usepackage{mathrsfs}
\usepackage{overpic}
\usepackage{booktabs}
\usepackage{pifont}
\usepackage{algorithm}
\usepackage{algorithmic}
\usepackage{balance}
\graphicspath{{./Imgs/}}

\usepackage{colortbl}
\usepackage{color,xcolor}

\usepackage{makecell}

\def\cvprPaperID{9854} 
\def\confName{CVPR}
\def\confYear{2023}

\begin{document}

\title{Fluxamba: Topology-Aware Anisotropic State Space Models for Geological Lineament Segmentation in Multi-Source Remote Sensing}

\author{
Jin Bai$^{1,2}$ \quad
Huiyao Zhang$^{1,2}$ \quad
Qi Wen$^{1}$\thanks{Corresponding author (email: wenqi@csu.ac.cn). This work was supported by the National Key Research and Development Program of China under Grant 2023YFB3906102.} \quad
Shengyang Li$^{1}$ \quad
Xiaolin Tian$^{3}$ \quad
Atta ur Rahman$^{4}$ \\
$^1$Technology and Engineering Center for Space Utilization, CAS, China \\
$^2$University of Chinese Academy of Sciences, China \\
$^3$Macau University of Science and Technology, China \\
$^4$University of Peshawar, Pakistan \\
{\tt\small baijin25@mails.ucas.ac.cn, zhanghuiyao25@csu.ac.cn, \{wenqi, shyli\}@csu.ac.cn} \\
{\tt\small xltian@must.edu.mo, atta-ur-rehman@uop.edu.pk}
}

\maketitle


\begin{abstract}
\vspace{-10pt}
The precise segmentation of geological linear features, spanning from planetary lineaments to terrestrial fractures, demands capturing long-range dependencies across complex anisotropic topologies. Although State Space Models (SSMs) offer near-linear computational complexity, their dependence on rigid, axis-aligned scanning trajectories induces a fundamental topological mismatch with curvilinear targets, resulting in fragmented context and feature erosion. To bridge this gap, we propose Fluxamba, a lightweight architecture that introduces a topology-aware feature rectification framework. Central to our design is the Structural Flux Block (SFB), which orchestrates an anisotropic information flux by integrating an Anisotropic Structural Gate (ASG) with a Prior-Modulated Flow (PMF). This mechanism decouples feature orientation from spatial location, dynamically gating context aggregation along the target's intrinsic geometry rather than rigid paths. Furthermore, to mitigate serialization-induced noise in low-contrast environments, we incorporate a Hierarchical Spatial Regulator (HSR) for multi-scale semantic alignment and a High-Fidelity Focus Unit (HFFU) to explicitly maximize the signal-to-noise ratio of faint features. Extensive experiments on diverse geological benchmarks—LROC-Lineament, LineaMapper, and GeoCrack—demonstrate that Fluxamba establishes a new state-of-the-art. Notably, on the challenging LROC-Lineament dataset, it achieves an F1-score of 89.22\% and mIoU of 89.87\%. Achieving a real-time inference speed of over 24 FPS with only 3.4M parameters and 6.3G FLOPs, Fluxamba reduces computational costs by up to two orders of magnitude compared to heavy-weight baselines, thereby establishing a new Pareto frontier between segmentation fidelity and onboard deployment feasibility.Our code is publicly available at \url{https://github.com/kbaijin/Fluxamba}. 
\end{abstract}

\section{Introduction}

Geological linear features, encompassing tectonic rilles on the Moon, cryovolcanic ridges on Europa, and fracture networks in terrestrial bedrock, serve as critical windows into planetary evolution and crustal dynamics \cite{Lineamapper, Geocrack, Lineament, watters2010evidence, kattenhorn2014evidence}. The precise extraction of these structures is paramount for applications ranging from autonomous rover navigation to geological hazard assessment \cite{verma2023autonomous}. However, unlike general visual objects with compact geometries, geological lineaments present a unique set of morphological challenges: they are inherently anisotropic, exhibiting extreme aspect ratios and intricate bifurcations, and often suffer from exceedingly low Signal-to-Noise Ratio (SNR) due to regolith erosion or complex textural interference \cite{mammoliti20233d, ahmadi2021fault, azarafza2021deep}.

Moreover, with the exponential growth of remote sensing data \cite{ma2015remote}, the traditional ``download-then-process'' paradigm faces severe bottlenecks due to limited downlink bandwidth \cite{furano2020towards}. Consequently, there is an urgent demand for real-time, onboard processing capabilities \cite{giuffrida2021varphi}. However, current deep learning paradigms face a fundamental dichotomy between representational capability and computational efficiency. 

\begin{figure}[t]
\centering
\includegraphics[width=1\columnwidth]{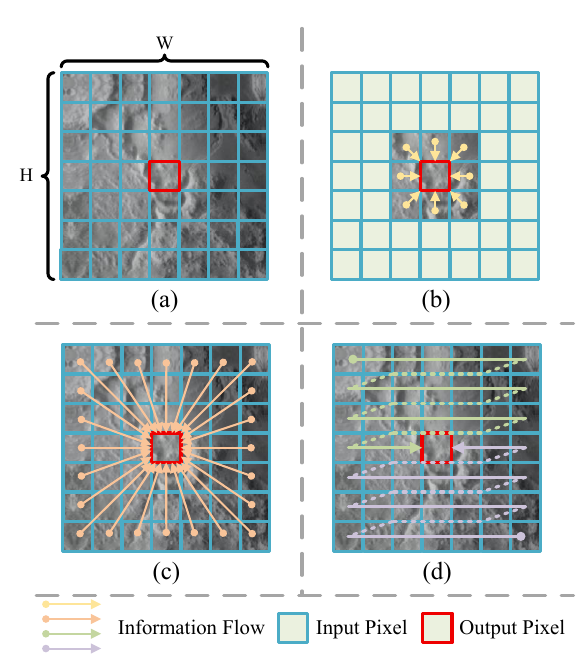} 
\caption{
Conceptual comparison of feature aggregation mechanisms across different architectures. (a) The input image representation with spatial resolution $H \times W$. (b) Convolution (CNNs): Aggregates features within a local receptive field, maintaining a linear computational complexity of $\mathcal{O}(HW)$. (c) Self-Attention (Transformers): Models global dependencies by densely calculating pairwise interactions between all pixels, incurring a quadratic complexity of $\mathcal{O}(H^2W^2)$. (d) State Space Models (SSMs): Recursively integrates context along specific scanning trajectories. While achieving near-linear complexity $\mathcal{O}(HW)$ similar to CNNs, it enables long-range dependency modeling with distance-dependent weight decay.
}
\label{figure1}
\end{figure}

As conceptually contrasted in Fig. \ref{figure1}(b)-(c), Convolutional Neural Networks (CNNs) and Vision Transformers (ViTs) represent opposing extremes of this spectrum.
CNNs, relying on local sliding windows (e.g., the seminal U-Net \cite{U-Net} and DeepCrack \cite{DeepCrack}), possess strong inductive biases. However, even with multi-scale enhancements like FPHBN \cite{FPHBN}, they are fundamentally constrained by limited Gaussian effective receptive fields \cite{wang2018non}, failing to model the long-range curvilinear dependencies essential for connecting disjoint geological segments.

Conversely, Vision Transformers (ViTs) \cite{VisionTransformer} circumvent this by utilizing self-attention mechanisms to establish dense, global pixel-to-pixel connections. Architectures such as Swin-UNet \cite{Swin-Unet} have demonstrated that modeling global pixel correlations significantly improves segmentation continuity. Yet, their quadratic complexity ($\mathcal{O}(N^2)$) incurs prohibitive memory and computational costs \cite{khan2022transformers}. This bottleneck renders ViTs impractical for deployment on resource-constrained edge platforms, such as satellite onboard processors or UAVs, where latency and power consumption are strictly bounded \cite{mehta2021mobilevit}.

\begin{figure}[t]
\centering
\includegraphics[width=1\columnwidth]{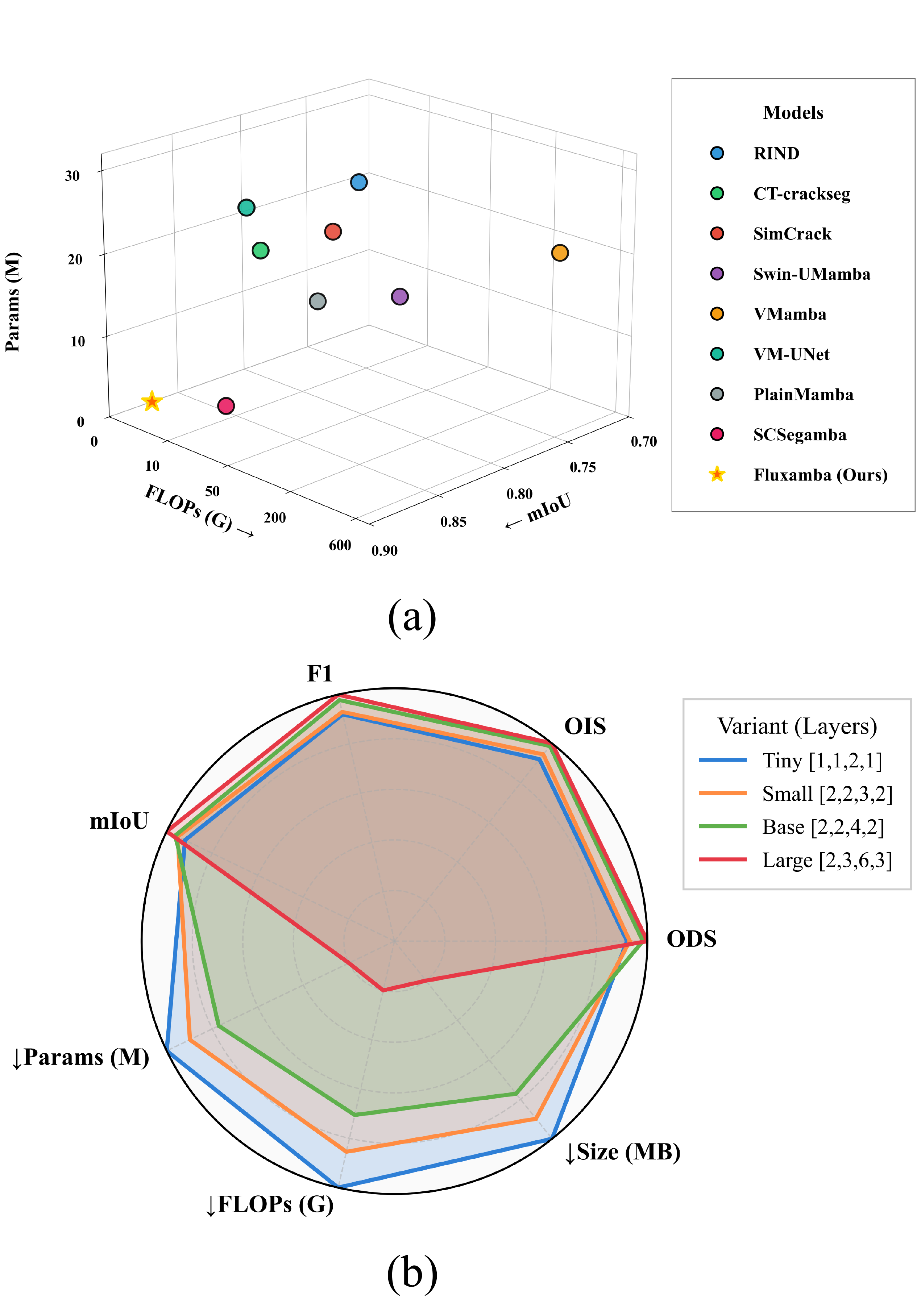} 
\caption{
Performance evaluation of the proposed Fluxamba. (a) Comparison with state-of-the-art (SOTA) methods in terms of parameter count, computational complexity (FLOPs), and segmentation accuracy (mIoU). Fluxamba (marked with a star) achieves a superior efficiency-accuracy trade-off. (b) Scalability analysis of Fluxamba variants (Tiny, Small, Base, Large) with different depth configurations. The radar chart illustrates the comprehensive impact of model scaling on computational cost (Size, FLOPs, Params) and segmentation metrics (F1, mIoU, OIS, ODS).
}
\label{figure2}
\end{figure}

Recently, State Space Models (SSMs) \cite{Mamba—architecture}, particularly Vision Mamba \cite{Vision-Mamba}, have attracted considerable attention for their ability to model long sequences with linear scaling characteristics ($\mathcal{O}(N)$) \cite{VMamba, zhang2024survey}. Despite this promise, we argue that while SSMs theoretically resolve the computational bottleneck, they introduce a new geometric bottleneck: a critical topological mismatch when applied to curvilinear geological lineaments.

As specifically illustrated in Fig. \ref{figure1}(d), existing methods rely on content-agnostic, rigid scanning trajectories (e.g., raster or snake scan) \cite{hu2024zigma, li2024mamba} to serialize 2D images. This imposes a disconnect between the 1D scanning sequence and the 2D curvilinear target: spatial neighbors along a tortuous crack may be scattered to distant time steps in the serialized sequence. Forcing anisotropic information flow along such predefined, axis-aligned paths inevitably severs semantic continuity and limits the model's ability to suppress off-axis background noise.

To resolve this conflict between rigid serialization and curvilinear topology, we propose \textbf{Fluxamba}, a lightweight architecture that introduces a novel paradigm of topology-aware information flux. Instead of relying on fixed scanning paths, Fluxamba employs a macro-level ``differentiable rectification'' strategy to dynamically align the effective information flux with the target's intrinsic geometry. Central to our design is the \textit{Structural Flux Block (SFB)}. Within this block, an \textit{Anisotropic Structural Gate (ASG)} acts as a topological probe to decouple feature orientation from spatial location. This geometric prior then guides the \textit{Prior-Modulated Flow (PMF)}, which adaptively gates and routes the information flux, creating a ``soft'' anisotropic pathway that adheres to the structural continuity of the lineament.

However, establishing geometric connectivity alone is insufficient in planetary scenarios where the signal is inherently weak. Complementing the geometric rectification, we further address the challenge of scale variance and signal degradation through a hierarchical filtering scheme. To bridge the inductive bias gap across encoder stages, we introduce a \textit{Hierarchical Spatial Regulator (HSR)}. By coordinating a \textit{Lightweight Modulation Refinement (LMR)} for shallow geometric details and a \textit{Global Transformer Reorganizer (GTR)} for deep semantic coherence, the HSR ensures robust feature alignment across resolutions. Finally, to strictly suppress regolith noise in low-contrast environments, we incorporate a \textit{High-Fidelity Focus Unit (HFFU)}. Adopting a non-residual, dual-polarized modulation strategy, the HFFU maximizes the SNR for faint features, ensuring that the topological connectivity established by the flux mechanism is preserved in the final output. 

The quantitative advantage of this architectural synergy is visually validated in Fig. \ref{figure2}, where Fluxamba establishes a superior efficiency-accuracy trade-off compared to SOTA baselines.

The main contributions of this article are summarized as follows:

\begin{itemize}
\item We propose Fluxamba, a lightweight, topology-aware State Space Model-based architecture tailored for geological linear feature segmentation. By replacing rigid scanning with a dynamic, rectification-based flux mechanism, it resolves the topological mismatch of standard SSMs. Compared with SOTA methods, Fluxamba maintains near-linear complexity while achieving superior boundary connectivity, making it highly suitable for future real-time application onboard.
\item We design the Structural Flux Block (SFB) to enforce anisotropic feature learning. By integrating the Anisotropic Structural Gate (ASG) and Prior-Modulated Flow (PMF), the SFB decouples feature propagation from rigid axes. This effectively fills the gap in existing SSMs regarding the modeling of irregular, tortuous geometric structures.
\item We introduce a robust signal enhancement strategy via the Hierarchical Spatial Regulator (HSR) and High-Fidelity Focus Unit (HFFU). By unifying the Lightweight Modulation Refinement (LMR) and Global Transformer Reorganizer (GTR), the HSR achieves scale-adaptive alignment, while the HFFU's innovative ``pure filtering'' strategy rigorously suppresses background noise in low-contrast environments (e.g., lunar regolith).
\item Fluxamba establishes a new state-of-the-art on three diverse benchmarks: LROC-Lineament, LineaMapper, and GeoCrack. Notably, it achieves an F1-score of 89.22\% on the challenging LROC-Lineament dataset with only 3.4M parameters, establishing a new Pareto frontier between segmentation fidelity and computational efficiency.
\end{itemize}

The remainder of this article is organized as follows. Section II reviews relevant literature. Section III details the methodology of the proposed Fluxamba architecture. Section IV presents experimental comparisons and analyses. Section V discusses generalization and onboard feasibility. Finally, Section VI concludes this article.

\section{Related Works and Motivations}

In this section, we critically review the literature focusing on three pivotal research streams that directly motivate our work: 1) specialized paradigms for anisotropic feature learning, which seek to transcend the limitations of isotropic convolutions; 2) the evolution of State Space Models (SSMs), where we identify a critical topological gap in current scanning strategies; and 3) signal enhancement methodologies in low-contrast environments, highlighting the limitations of residual-based learning for faint feature extraction.

\subsection{Anisotropic Feature Learning and Geometric Adaptation}
Distinct from general semantic segmentation objects with compact shapes, geological lineaments are characterized by extreme aspect ratios and complex, curvilinear topologies. This morphological uniqueness necessitates network designs that possess strong geometric inductive biases, going beyond standard isotropic receptive fields.
\vspace{-10pt}
\paragraph{Boundary-Aware Constraints.}
To address the challenge of boundary blurring inherent in encoder-decoder architectures, recent research has explicitly incorporated boundary-aware mechanisms. BARNet \cite{BARNet} explicitly models boundary gradients to sharpen fracture edges, while RIND \cite{RINDNet} utilizes edge detection constraints to refine segmentation outputs for discontinuities in reflectance and depth. Similarly, ADDUNet \cite{ADDUNet} integrates attention gates to capture fine characteristics. While these methods improve local edge precision, they typically rely on CNN backbones. The limited Gaussian effective receptive fields of CNNs often fail to model the long-range dependencies necessary to connect disjoint segments of elongated structures, leading to topological fragmentation.
\vspace{-10pt}
\paragraph{Long-Range Anisotropic Aggregation.}
To accommodate curvilinear topologies, contemporary research has sought to break the fixed-grid limitation. Strategies such as Strip Pooling \cite{hou2020strip} and its remote sensing adaptations \cite{cao2022csanet} employ anisotropic pooling kernels to capture long-range axial dependencies. Concurrently, Deformable Convolutional Networks (DCNs) \cite{dai2017deformable} and active contour models \cite{peng2020deep} introduce geometric adaptivity by dynamically adjusting spatial sampling locations based on local feature content.

However, a fundamental limitation persists in these data-driven deformation paradigms. DCNs rely on the \textit{implicit} learning of offsets driven solely by local gradients. In signal-starved planetary environments, this unconstrained local deformation renders the model susceptible to background textural interference, leading to ``semantic drift'' where sampling points diverge from the true fracture trajectory. This limitation necessitates mechanisms that not only adapt but explicitly \textit{rectify} the information flux via geometric priors, rather than relying solely on implicit data-driven deformations.

\subsection{State Space Models and the Topological Gap}
State Space Models (SSMs), particularly the Mamba architecture \cite{Mamba—architecture}, have emerged as a formidable alternative to Transformers, offering a promising middle ground by modeling global context with linear computational complexity ($\mathcal{O}(N)$) via a Selective Scan (S6) mechanism.
\vspace{-10pt}
\paragraph{Serialization in Vision SSMs.}
Adapting the causal 1D nature of Mamba to 2D vision has spurred specialized serialization strategies. Vision Mamba (Vim) \cite{Vision-Mamba} employs bidirectional scanning, while VMamba \cite{VMamba} proposes a Cross-Scan Module (CSM) to traverse spatial features in four cardinal directions. Hybrid approaches like Swin-UMamba \cite{Swin-UMamba} and VM-UNet \cite{VM-UNet} further integrate Mamba blocks into hierarchical U-Net structures. Concurrently, the remote sensing community has adopted Mamba for tasks such as scene classification (RSMamba \cite{RSMamba}), pansharpening (Pan-Mamba \cite{Pan-Mamba}), and multi-source fusion (FusionMamba \cite{FusionMamba}).
\vspace{-10pt}
\paragraph{The Problem of Rigid Scanning.}
Most relevant to our work, the recent SCSegamba \cite{SCSegamba} proposes a Structure-Aware Scanning Strategy (SASS) tailored for structural defects. While SCSegamba attempts to mitigate topological disconnection by incorporating specialized scanning templates (e.g., diagonal snake paths), it remains fundamentally bound by discrete, predefined trajectories. We argue that simply augmenting the set of rigid scanning patterns acts merely as a \textit{piecewise approximation}, which still struggles to align with the continuous, arbitrary curvature of geological lineaments (a phenomenon we define as ``spatial coherence dilution''). This highlights the need for a paradigm shift from rigid traversal to differentiable rectification strategies, capable of synthesizing continuous, topology-aware flows that strictly adhere to the intrinsic geometry of the target.

\subsection{Signal Enhancement in Low-Contrast Regimes}
Geological lineaments on planetary surfaces often suffer from exceedingly low Signal-to-Noise Ratio (SNR) due to regolith erosion, variable illumination, and complex background textures. This presents a unique challenge: distinguishing faint structural signals from high-frequency noise (e.g., speckle or crater rims).
\vspace{-10pt}
\paragraph{Limitations of Residual Learning.}
To enhance feature representation, standard approaches widely employ Attention Mechanisms (e.g., SE, CBAM) \cite{hu2018squeeze, woo2018cbam} or multi-scale fusion modules \cite{lin2017feature}. However, the dominant paradigm in these modules is Residual Learning ($y = x + \mathcal{F}(x)$) \cite{he2016deep}. While residual connections are essential for gradient propagation in deep networks, we argue that they are suboptimal for low-contrast geological segmentation. The identity mapping ($x$) indiscriminately preserves both the weak target signal and the dominant background noise. In scenarios where noise variance exceeds signal strength, the residual addition inadvertently reinforces background artifacts, hindering the segmentation of faint targets \cite{zhang2017beyond, li2019selective}.
\vspace{-10pt}
\paragraph{Necessity for Pure Filtering.}
This observation motivates a shift from ``feature preservation'' to ``noise rejection.'' Consequently, in low-contrast regimes, there is a critical need for specialized filtering mechanisms that depart from standard residual paradigms to strictly suppress background variance without the interference of an identity path. Furthermore, addressing the scale variance of lineaments masked by noise requires hierarchical alignment strategies to ensure that semantic enhancement does not come at the cost of spatial precision.

\begin{figure*}[tb]
\centering
\includegraphics[width=1\textwidth]{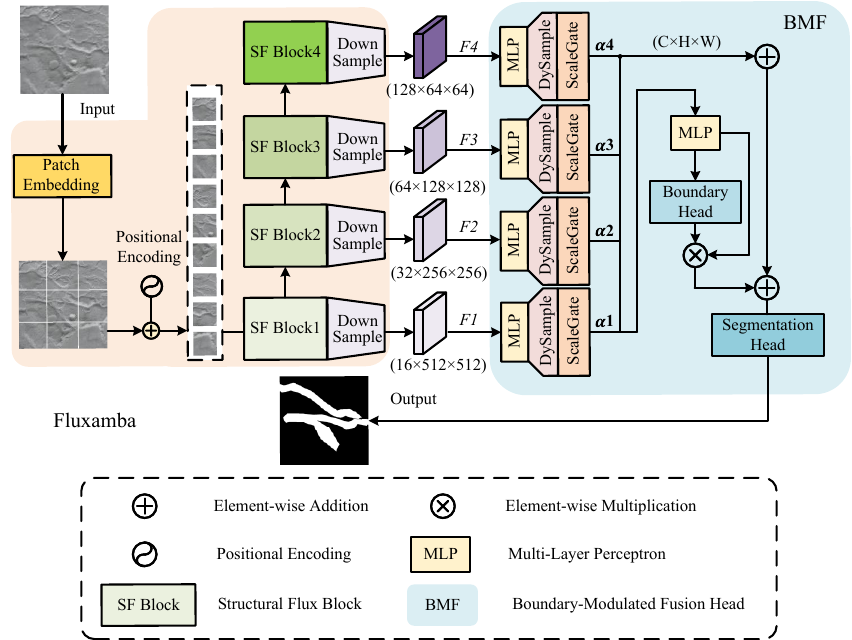} 
\caption{
Overall architecture of the proposed Fluxamba. The framework adopts a hierarchical encoder-decoder structure. 
On the left, the encoder utilizes stacked Structural Flux Blocks (SF Blocks) to extract multi-scale anisotropic features, denoted as $F_1$ to $F_4$, at varying resolutions through downsampling. 
On the right, the decoder incorporates the Boundary-Modulated Fusion (BMF) module, which dynamically aggregates these hierarchical features via DySample and ScaleGate operations using element-wise addition ($\oplus$). 
To enhance linear structural details, a specialized boundary head composed of a Conv-BN-ReLU sequence and a $1\times1$ convolution modulates the fused features via element-wise multiplication ($\otimes$). 
Finally, the segmentation head comprising Conv-BN-ReLU, dropout, and a $1\times1$ convolution layer generates the prediction map.
}
\label{fig:overall_arch}
\end{figure*}

\begin{figure*}[tb]
\centering
\includegraphics[width=1\textwidth]{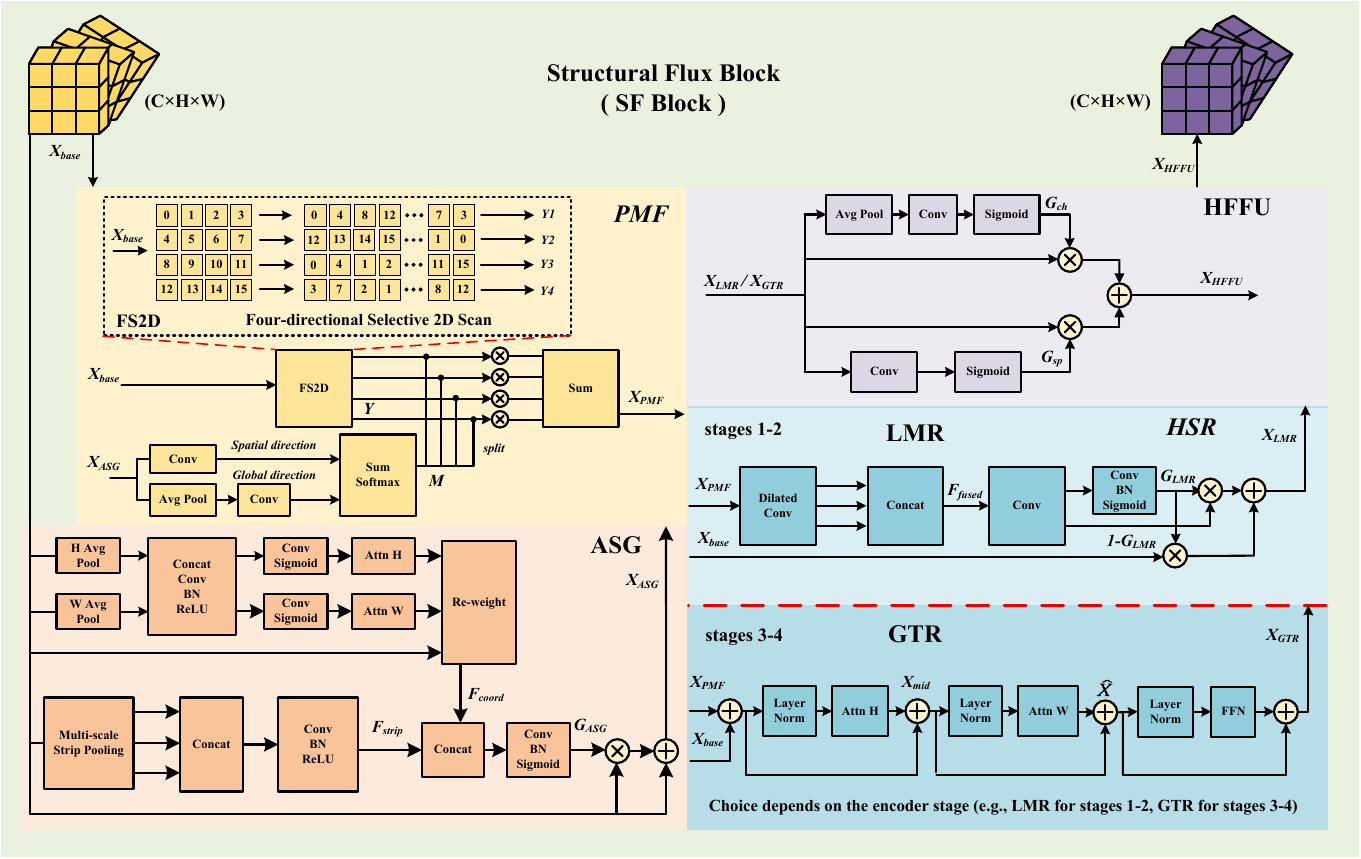}
\caption{
Detailed architecture of the proposed Structural Flux Block (SFB). Designed as the core computational unit, the SFB orchestrates a topology-aware information flux through four synergistic stages: (1) Anisotropic Structural Gate (ASG): Extracts geometric priors to model long-range dependencies; (2) Prior-Modulated Flow (PMF): Rectifies the rigid FS2D scanning trajectories by dynamically gating the information flow based on ASG priors; (3) Hierarchical Spatial Regulator (HSR): Aligns feature semantics adaptively, utilizing Lightweight Modulation Refinement (LMR) for local detail preservation in shallow layers (Stages 1-2) and Global Transformer Reorganizer (GTR) for semantic coherence in deep layers (Stages 3-4); (4) High-Fidelity Focus Unit (HFFU): Maximizes the signal-to-noise ratio via dual-polarized modulation.
}
\label{fig:sfb_structure}
\end{figure*}

\section{Methodology}

\subsection{Overall Architecture}
The proposed Fluxamba establishes a hierarchical encoder-decoder framework specifically tailored for the high-fidelity segmentation of geological linear structures, as illustrated in Fig. \ref{fig:overall_arch}. To address the intrinsic topological discrepancy between the rigid, axis-aligned scanning inherent in standard State Space Models (SSMs) and the curvilinear nature of geological targets, we introduce a fundamental computational unit designated as the Structural Flux Block (SFB).
The architecture comprises a four-stage encoder that utilizes stacked SFBs to extract multi-scale anisotropic features ($\mathbf{F}_1$ to $\mathbf{F}_4$) at varying resolutions. To transform these encoded representations into a precise segmentation map, the decoder incorporates a Boundary-Modulated Fusion (BMF) module. As shown in the decoder section of Fig. \ref{fig:overall_arch}, the BMF dynamically aggregates hierarchical features via content-aware upsampling and ScaleGate selection, while explicitly leveraging a boundary head to modulate the fused features, thereby ensuring the structural integrity of the final prediction.

\subsection{Structural Flux Block (SFB)}
The SFB functions as the fundamental unit of Fluxamba. Rather than altering the standard linear-complexity scanning trajectory—a modification that would inevitably compromise inference efficiency—the SFB employs a novel "differentiable rectification" mechanism. As detailed in Fig. \ref{fig:sfb_structure}, it coordinates a topology-aware information flux through four synergistic components: the Anisotropic Structural Gate (ASG) for geometric prior extraction, the Prior-Modulated Flow (PMF) for trajectory rectification, the Hierarchical Spatial Regulator (HSR) for scale-adaptive alignment, and the High-Fidelity Focus Unit (HFFU) for noise suppression.

\begin{figure*}[tb]
\centering
\includegraphics[width=1\textwidth]{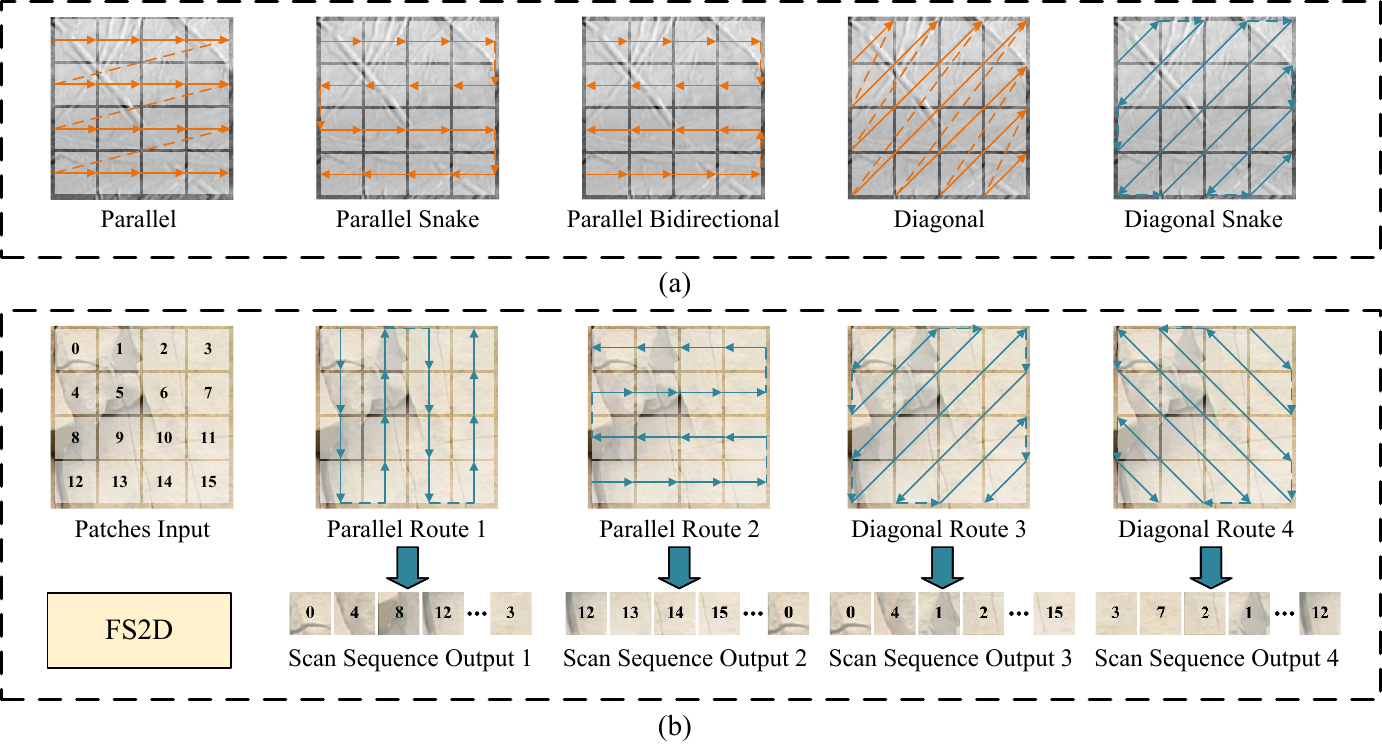}
\caption{
Illustration of scanning primitives and the sequence generation pipeline. (a) Visualization of candidate scanning trajectories: diverse path patterns (e.g., Parallel, Diagonal) utilized to capture spatial context from varying orientations. (b) Serialization execution flow: The process of transforming 2D patches into direction-specific 1D sequences. These multi-route sequences provide the necessary anisotropic inputs for the subsequent Prior-Modulated Flow (PMF), allowing the model to dynamically gate and re-weight different strategies based on the target structure.
}
\label{fig:scanning}
\end{figure*}

\subsection{Anisotropic Structural Gate (ASG)}
The Anisotropic Structural Gate (ASG) functions as the topological encoder within the SFB, explicitly designed to model the long-range dependencies and complex geometries inherent in geological linear targets.
As depicted in the ASG panel of Fig. \ref{fig:sfb_structure}, the module processes the backbone input $\mathbf{X}_{\text{base}} \in \mathbb{R}^{C \times H \times W}$ through a dual-branch architecture that decouples spatial encoding from contextual aggregation. The Coordinate-Aware Branch \cite{hou2021coordinate} performs axis-separated average pooling to generate $\mathbf{F}_{\text{coord}}$, effectively encoding precise spatial positioning information along horizontal and vertical directions. In parallel, the Strip-Pooling Branch \cite{hou2020strip} employs multi-scale strip pooling kernels to break the bounded receptive field of conventional convolutions, allowing the network to aggregate anisotropic context along narrow, elongated trajectories ($\mathbf{F}_{\text{strip}}$).
These position-sensitive and shape-aware cues are spatially broadcasted and synergistically fused to synthesize the structural gate $\mathbf{G}_{\text{ASG}}$:
\begin{equation}
\mathbf{G}_{\text{ASG}} = \sigma \left( \mathcal{C}_{1\times1} \left( \operatorname{Cat}(\mathbf{F}_{\text{coord}}, \mathbf{F}_{\text{strip}}) \right) \right)
\end{equation}
where $\operatorname{Cat}(\cdot)$ denotes channel-wise concatenation, $\mathcal{C}_{1\times1}$ serves as a channel projection layer, and $\sigma$ represents the Sigmoid activation. Finally, this gate dynamically modulates the input feature via a residual attention mechanism:
\begin{equation}
\mathbf{X}_{\text{ASG}} = \mathbf{X}_{\text{base}} + \mathbf{X}_{\text{base}} \odot \mathbf{G}_{\text{ASG}}
\end{equation}
where $\odot$ denotes the element-wise Hadamard product. This yields the geometry-enhanced representation $\mathbf{X}_{\text{ASG}}$.

\subsection{Prior-Modulated Flow (PMF)}
The Prior-Modulated Flow (PMF) addresses the inherent limitation of rigid scanning trajectories via a novel feature rectification strategy. While standard Vision Mamba models \cite{Vision-Mamba} rely on fixed, axis-aligned scanning paths (as visualized in Fig. \ref{fig:scanning}(a)), which often fragment continuous linear features due to a lack of geometric adaptability, the proposed PMF overcomes this not by altering the underlying micro-level SSM equations, but by introducing a \textit{macro-level} rectification layer. This design allows us to retain the computationally efficient Four-directional Selective 2D Scan (FS2D) for token serialization while effectively emulating a topology-aware information flux through post-scan aggregation.

The module operates through two parallel streams. The Primary Stream processes the backbone features $\mathbf{X}_{\text{base}}$ using the standard FS2D mechanism—whose serialization logic is detailed in Fig. \ref{fig:scanning}(b)—generating four independent directional state sequences $\{\mathbf{Y}_k\}_{k=1}^4$.

As illustrated in the PMF diagram in Fig. \ref{fig:sfb_structure}, the Modulation Stream constructs a dynamic weight map $\mathbf{M}$ conditioned on the structural prior $\mathbf{X}_{\text{ASG}}$ to rectify the uniform aggregation of these sequences. We first aggregate multi-scale contexts to compute the raw modulation weights:
\begin{equation}
\mathbf{M} = \mathcal{C}_{\text{local}}\left(\mathbf{X}_{\text{ASG}}\right) + \mathcal{C}_{\text{global}}\left(\operatorname{GAP}(\mathbf{X}_{\text{ASG}})\right)
\end{equation}
where $\mathcal{C}_{\text{local}}$ and $\mathcal{C}_{\text{global}}$ denote convolutional projection layers, and $\operatorname{GAP}$ represents global average pooling. The globally pooled feature is spatially broadcasted to match the resolution of the local branch before addition. Subsequently, $\mathbf{M}$ is partitioned and normalized via a Split-Softmax operation to yield four directional sub-gates $\{\mathbf{M}_k\}_{k=1}^4$. These sub-gates function as differentiable "soft switches," establishing a competitive mechanism where the network selectively amplifies signal propagation along paths that align with the target's intrinsic curvature while suppressing misalignment noise.
The final output $\mathbf{X}_{\text{PMF}}$ is obtained by the weighted aggregation of the rectified flows:
\begin{equation}
\mathbf{X}_{\text{PMF}} = \sum_{k=1}^{4} (\mathbf{Y}_k \odot \mathbf{M}_k)
\end{equation}
By dynamically recalibrating the contribution of each scanning route, this mechanism synthesizes "virtual" anisotropic pathways that strictly follow the target's structural continuity, effectively bridging the gap between discrete rigid scanning and continuous geological topology.

\textit{Geometric Interpretation of Rectification:} It is noteworthy that PMF achieves rotational robustness logically via the learned weight map $\mathbf{M}$, rather than physically rearranging the pixel scanning sequence. By treating the four cardinal scanning directions ($0^\circ, 45^\circ, 90^\circ, 135^\circ$) as a set of geometric basis vectors, the network learns to synthesize omnidirectional feature representations through the weighted superposition of these bases. This linear combination of cardinal directions effectively enables rotational robustness analogous to continuous approaches, allowing Fluxamba to model geological features of arbitrary orientations while preserving the hardware-friendly parallelization of standard SSMs.

\subsection{Hierarchical Spatial Regulator (HSR)}
To mitigate the inductive bias gap between the rectified features $\mathbf{X}_{\text{PMF}}$ and the original backbone representation $\mathbf{X}_{\text{base}}$, the Hierarchical Spatial Regulator (HSR) adaptively aligns feature semantics according to the encoder stage depth $s$ (detailed in the HSR panel of Fig. \ref{fig:sfb_structure}). This design exploits the observation that shallow stages encode high-frequency geometric details, while deep stages prioritize abstract semantic coherence.

Geometry Preservation in Shallow Stages ($s \in \{1, 2\}$).
In early stages, where preserving local boundary precision is paramount, we employ \textit{Lightweight Modulation Refinement (LMR)}. LMR synthesizes a multi-scale structural representation by aggregating context from diverse receptive fields via parallel depthwise dilated convolutions \cite{yu2015multi}.
Let $\mathcal{D}_r$ denote a depthwise convolution with dilation rate $r$, and $\mathcal{R} = \{r_1, \dots, r_n\}$ be the set of dilation rates. We extract and concatenate multi-granular features as follows:
\begin{equation}
\mathbf{F}_{\text{cat}} = \operatorname{Cat}\left( \left\{ \mathcal{D}_{r}(\mathbf{X}_{\text{PMF}}) \right\}_{r \in \mathcal{R}} \right)
\end{equation}
To align dimensions, $\mathbf{F}_{\text{cat}}$ is projected into a unified feature space $\tilde{\mathbf{F}} = \mathcal{C}_{\text{proj}}(\mathbf{F}_{\text{cat}})$. Subsequently, a gating mechanism generates the modulation mask $\mathbf{G}_{\text{LMR}}$:
\begin{equation}
\mathbf{G}_{\text{LMR}} = \sigma\left( \mathcal{C}_{\text{gate}} \left( \tilde{\mathbf{F}} \right) \right)
\end{equation}
The final output $\mathbf{X}_{\text{LMR}}$ is formed via a gated injection scheme, selectively incorporating the refined multi-scale context into the base manifold:
\begin{equation}
\mathbf{X}_{\text{LMR}} = (1 - \mathbf{G}_{\text{LMR}}) \odot \mathbf{X}_{\text{base}} + \mathbf{G}_{\text{LMR}} \odot \tilde{\mathbf{F}}
\end{equation}

Semantic Reorganization in Deep Stages ($s \in \{3, 4\}$).
Deep stages demand long-range contextual consistency. Here, we utilize the \textit{Global Transformer Reorganizer (GTR)} to model global dependencies. GTR first fuses the rectified flow with the base feature via a residual connection to establish a comprehensive initialization that retains original spatial gradients:
\begin{equation}
\mathbf{X}_{\text{in}} = \mathbf{X}_{\text{PMF}} + \mathbf{X}_{\text{base}}
\end{equation}
To maintain computational efficiency while capturing global context, GTR applies sequential axial attention \cite{ho2019axial} rather than computationally prohibitive full self-attention \cite{vaswani2017attention}. The feature $\mathbf{X}_{\text{in}}$ is processed sequentially by vertical ($\operatorname{MSA}_H$) and horizontal ($\operatorname{MSA}_W$) attention modules:
\begin{equation}
\begin{split}
\mathbf{X}_{\text{H}} &= \operatorname{MSA}_H(\operatorname{LN}(\mathbf{X}_{\text{in}})) + \mathbf{X}_{\text{in}} \\
\mathbf{X}_{\text{W}} &= \operatorname{MSA}_W(\operatorname{LN}(\mathbf{X}_{\text{H}})) + \mathbf{X}_{\text{H}}
\end{split}
\end{equation}
where $\operatorname{LN}$ denotes Layer Normalization. Although axial attention theoretically entails a complexity of $\mathcal{O}(HW(H+W))$, the GTR is exclusively deployed in the deepest stages ($s \in \{3, 4\}$) where the spatial resolution is significantly downsampled. Consequently, the actual computational overhead is negligible compared to the linear complexity $\mathcal{O}(HW)$ dominated by the shallow layers. The globally reorganized feature is finally refined via a Feed-Forward Network (FFN):
\begin{equation}
\mathbf{X}_{\text{GTR}} = \operatorname{FFN}(\operatorname{LN}(\mathbf{X}_{\text{W}})) + \mathbf{X}_{\text{W}}
\end{equation}
Consequently, the HSR unifies these stage-specific paradigms: $\mathbf{X}_{\text{HSR}}^{(s)} = \mathbf{X}_{\text{LMR}}^{(s)}$ for $s \le 2$, and $\mathbf{X}_{\text{GTR}}^{(s)}$ for $s \ge 3$.

\subsection{High-Fidelity Focus Unit (HFFU)}
To counteract the cumulative serialization noise introduced by the preceding scanning operations, the High-Fidelity Focus Unit (HFFU) is designed to explicitly maximize the signal-to-noise ratio (SNR) via a non-residual Dual-Polarized Modulation scheme (visualized in the top-right panel of Fig. \ref{fig:sfb_structure}). Unlike standard residual attention modules that typically preserve background noise via identity connections, HFFU employs a pure filtering strategy. Building upon the structurally aligned features from the preceding HSR module, this unit functions as a highly selective soft gate, operating through two concurrent branches to isolate salient structural components:
\begin{equation}
\begin{split}
\mathbf{G}_{\text{ch}} &= \sigma\left(\mathcal{C}_{\text{excite}}\left(\operatorname{GAP}(\mathbf{X}_{\text{HSR}})\right)\right) \\
\mathbf{G}_{\text{sp}} &= \sigma\left(\mathcal{C}_{\text{spatial}}\left(\mathbf{X}_{\text{HSR}}\right)\right)
\end{split}
\end{equation}
where $\mathcal{C}_{\text{excite}}$ and $\mathcal{C}_{\text{spatial}}$ denote the channel excitation and spatial projection layers, respectively. The Channel Polarization Branch ($\mathbf{G}_{\text{ch}}$) suppresses irrelevant feature channels, while the Spatial Polarization Branch ($\mathbf{G}_{\text{sp}}$) highlights topological connectivity. The final rectified output $\mathbf{X}_{\text{HFFU}}$ is obtained via the additive aggregation of these filtered streams:
\begin{equation}
\mathbf{X}_{\text{HFFU}} = \left( \mathbf{G}_{\text{ch}} \odot \mathbf{X}_{\text{HSR}} \right) + \left( \mathbf{G}_{\text{sp}} \odot \mathbf{X}_{\text{HSR}} \right)
\end{equation}
This design ensures that artifacts and non-structural background regions are effectively attenuated before the features propagate to the next stage. By deliberately omitting the residual connection in this final stage, HFFU acts as a 'hard' spectral gate, forcing the network to discard low-confidence background signals that would otherwise propagate through an identity path.Although removing residual connections typically risks optimization instability, we empirically demonstrate in Section V.A (Fig. 11) that our architecture maintains robust convergence properties.

\subsection{Boundary-Modulated Fusion (BMF) Head}
The BMF Head, illustrated in the decoder section of Fig. \ref{fig:overall_arch}, ensures precise delineation by adaptively integrating multi-scale features $\{\mathbf{F}_s\}_{s=1}^4$ and explicitly enforcing structural fidelity.

To transcend the inherent limitations of standard bilinear interpolation, we devise a content-aware aggregation strategy that synergizes feature alignment with semantic recalibration. We first unify the resolution of hierarchical features using the dynamic upsampling operator $\mathcal{F}_{\text{up}}$ \cite{liu2023learning}, and subsequently re-weight them via a ScaleGate mechanism adapted from \cite{shi2023transformer}. The aligned and recalibrated feature $\mathbf{F}'_s$ for stage $s$ is computed as:
\begin{equation}
\mathbf{F}'_s = \boldsymbol{\alpha}_s \odot \mathcal{F}_{\text{up}}(\operatorname{MLP}(\mathbf{F}_s)), \quad s \in \{1, \dots, 4\}
\end{equation}
where $\odot$ denotes the element-wise Hadamard product. Here, $\boldsymbol{\alpha}_s \in \mathbb{R}^C$ represents the learnable channel-wise gating vector derived via global context modeling \cite{shi2023transformer}, while $\operatorname{MLP}$ performs channel mixing. We then aggregate these optimized contexts via summation ($\mathbf{F}_{\Sigma} = \sum \mathbf{F}'_s$).

Crucially, to counteract the erosion of fine-grained boundary details—a common degradation in deep supervision—we employ a Boundary-Aware Residual Injection strategy. Leveraging the high-frequency spatial information preserved in the shallowest stage, an auxiliary head predicts a probabilistic boundary map $\mathbf{M}_{\text{bound}}$ from the aligned Stage-1 feature $\mathbf{F}'_1$:
\begin{equation}
\mathbf{M}_{\text{bound}} = \sigma(\mathcal{H}_{\text{bound}}(\mathbf{F}'_1))
\end{equation}
This map functions as a structural attention mask. Rather than indiscriminately modulating the fused features, it selectively highlights high-frequency boundaries within the projected Stage-1 feature map ($F'_1$). This boundary-enhanced signal is then injected into the global context ($F_{\Sigma}$) via a residual connection, ensuring that geometric details are preserved without being diluted by deep semantic abstractions:
\begin{equation}
\mathbf{F}_{\text{fused}} = \mathbf{F}_{\Sigma} + \lambda \cdot \left( \mathcal{M}_{\text{proj}}(\mathbf{F}'_1) \odot \mathbf{M}_{\text{bound}} \right)
\end{equation}
where $\mathcal{M}_{\text{proj}}$ denotes the projection MLP and $\lambda$ is a balancing factor (defaulting to 1.0). Finally, the high-fidelity segmentation output $\mathbf{Y}_{\text{output}}$ is generated by projecting this geometrically refined representation:
\begin{equation}
\mathbf{Y}_{\text{output}} = \mathcal{H}_{\text{seg}}(\mathbf{F}_{\text{fused}})
\end{equation}
This formulation ensures that faint linear structures are explicitly sharpened by the boundary prior immediately before the final prediction.

\subsection{Objective Function}
\label{subsec:objective_function}

To mitigate the extreme foreground-background imbalance characteristic of planetary lineaments and enhance the delineation of faint structures, we employ a hybrid objective function. This objective integrates pixel-level weighted supervision, global structural constraints, and explicit boundary refinement, formulated as:

\begin{equation}
\label{eqn:total_loss}
\mathcal{L}_{\text{total}} = \lambda_{\text{bce}} \mathcal{L}_{\text{WBCE}} + \lambda_{\text{dice}} \mathcal{L}_{\text{Dice}} + \lambda_{b} \mathcal{L}_{\text{boundary}}
\end{equation}

The Weighted Binary Cross-Entropy ($\mathcal{L}_{\text{WBCE}}$) maintains pixel-level fidelity while counteracting background dominance by enforcing a positive class weight $w_{pos}$ \cite{xie2015holistically}:
\begin{equation}
\label{eqn:wbce}
\small 
\mathcal{L}_{\text{WBCE}} = -\frac{1}{N} \sum_{i=1}^{N} [w_{pos} y_i \log(p_i) + (1-y_i) \log(1-p_i)]
\end{equation}
where $y_i \in \{0, 1\}$ and $p_i$ denote the ground truth and predicted probability, respectively. To ensure global topological continuity, we incorporate the Soft Dice Loss ($\mathcal{L}_{\text{Dice}}$) \cite{milletari2016v} to maximize region-based overlap:
\begin{equation}
\label{eqn:dice}
\mathcal{L}_{\text{Dice}} = 1 - \frac{2 \sum_{i=1}^{N} p_i y_i + \epsilon}{\sum_{i=1}^{N} p_i + \sum_{i=1}^{N} y_i + \epsilon}
\end{equation}
where $\epsilon$ is a smoothing term. Furthermore, to sharpen curvilinear edges, $\mathcal{L}_{\text{boundary}}$ supervises the auxiliary boundary head \cite{qin2019basnet}. The ground truth boundaries are dynamically generated via morphological gradients (dilation minus erosion) on the segmentation masks.

In our implementation, hyperparameters are empirically set to $\lambda_{\text{bce}}=0.3$, $\lambda_{\text{dice}}=0.4$, and $\lambda_{b}=0.2$. Crucially, for sparse geological targets, we set $w_{pos}=5.0$ to rigorously penalize false negatives.

\section{Experiments}
\subsection{Datasets}
To comprehensively evaluate the topology-aware modeling capabilities of Fluxamba across varying planetary and terrestrial environments, we establish a diverse benchmark suite comprising three geological datasets: LROC-Lineament \cite{Lineament}, LineaMapper \cite{Lineamapper}, and GeoCrack \cite{Geocrack}. These datasets are strategically selected to represent a broad spectrum of segmentation challenges, ranging from low-contrast planetary lineaments to intricate terrestrial fracture networks.
\vspace{-10pt}
\paragraph{LROC-Lineament.} 
Derived from the Lunar Reconnaissance Orbiter Camera (LROC) archive, this dataset targets the global continuity of anisotropic structures on the lunar surface. It comprises 1,000 high-resolution images ($100$ m/pixel) encompassing four primary linear categories: ridges, rills, catenae, and valles. Ground truth labels were generated via rigorous pixel-level annotation utilizing the Lunar Orbital Data Explorer. Following standard protocols, the dataset is partitioned into 800 training, 100 validation, and 100 testing samples.
The primary challenge of LROC-Lineament lies in the extremely low Signal-to-Noise Ratio (SNR) and ambiguous boundary definitions induced by impact cratering and variable illumination. This serves as a rigorous testbed for the model's ability to preserve structural integrity and suppress background noise in signal-starved environments.
\vspace{-10pt}
\paragraph{LineaMapper.} 
To evaluate segmentation performance on complex icy bodies, we utilize the LineaMapper dataset derived from the Galileo Solid-State Imager (SSI) mosaics of Europa. This dataset consists of approximately 930 tiles extracted from 15 global mosaics, capturing four distinct cryovolcanic tectonic units: bands, double ridges, ridge complexes, and undifferentiated lineae.
Unlike lunar features, Europa's surface is characterized by intricate cross-cutting relationships and heterogeneous imaging quality. This dataset benchmarks the model's capacity to disentangle overlapping topological instances and maintain semantic consistency across fractured ice shells—a task where traditional rigid scanning often fails due to the tortuous and bifurcated geometry of ridge complexes.
\vspace{-10pt}
\paragraph{GeoCrack.} 
Representing terrestrial geological complexity, GeoCrack is a high-fidelity dataset of natural rock fractures collected from 11 geologically distinct sites across Europe and the Middle East (e.g., carbonate platforms in Italy and ophiolites in Oman). Data acquisition employed both terrestrial photography and Unmanned Aerial Vehicle (UAV) surveys to ensure multi-scale coverage. The dataset encompasses a rich variety of host lithologies, including Cretaceous limestones, dolomites, and ultrabasic peridotites.
GeoCrack poses a unique challenge due to the intense textural interference from natural rock surfaces and the irregular, multi-scale morphology of natural fractures. It serves to validate Fluxamba's robustness in distinguishing micro-scale fracture boundaries from complex background textures.

\begin{table*}[htb]
    \caption{Comparison with 8 SOTA methods across 3 datasets. All values are in percentage (\%). Best results are in bold, and second-best results are underlined.}
    \centering
    \setlength\tabcolsep{1mm} 
    \fontsize{8pt}{10pt}\selectfont 
    
    \begin{tabular}{l|cccccc|cccccc|cccccc} 
        \hline 
        \multirow{2}{*}{Methods} & 
        \multicolumn{6}{c|}{GeoCrack (\%)} & 
        \multicolumn{6}{c|}{LineaMapper (\%)} &
        \multicolumn{6}{c}{LROC-Lineament (\%)} \\
        \cline{2-7} \cline{8-13} \cline{14-19}
        & ODS & OIS & P & R & F1 & mIoU & ODS & OIS & P & R & F1 & mIoU & ODS & OIS & P & R & F1 & mIoU \\
        \hline 
        RIND \cite{RINDNet} & 79.49 & 78.39 & 78.94 & 80.04 & 79.49 & 75.26 & 82.44 & 80.19 & 81.02 & 83.91 &  82.44 & 82.64 & 80.76 & 80.08 & 74.93 & 83.37 & 77.58 & 82.40 \\
        CT-crackseg \cite{CT-CrackSeg} & 81.59 & 80.86 & 82.81 & \underline{80.41} & 81.59 & 78.14 & 85.82 & 84.38 & 84.02 & 87.53 & 85.82 & 85.06 & 68.92 & 72.03 & 63.01 & 78.75 & 70.01 & 86.30 \\
        SimCrack \cite{SimCrack} & 82.16 & \underline{81.98} & \underline{84.78} & 79.69 & 82.16 & 79.16 & 86.07 & 84.51 & 84.17 & 87.38 & 86.07 & \underline{85.48} & 86.72 & \underline{85.46} & 87.61 & \underline{86.47} & \underline{86.38} & \underline{88.88} \\
        Swin-UMamba \cite{Swin-UMamba} & 70.79 & 69.57 & 69.39 & 72.25 & 70.79 & 66.47 & 83.43 & 81.49 & 81.64 &  85.34 & 83.43 & 83.69 & 85.99 & 85.03 & 87.29 & 82.08 & 83.48 & 87.08 \\
        VMamba \cite{VMamba} & 76.92 & 75.66 & 79.91 & 74.15 & 76.92 & 74.18 & 85.12 & 82.85 & 83.72 & 86.58 & 85.12 & 83.36 & 64.45 & 62.07 & 62.85 & 60.14 & 59.33 & 72.11 \\
        VM-UNet \cite{VM-UNet} & \underline{82.71} & 81.62 & 84.51 & 79.42 & \underline{82.71} & \underline{79.75} & \underline{86.31} & \underline{84.72} & \underline{84.76} & \underline{87.76} & \underline{86.31} & 84.91 & \underline{87.28} & 84.59 & \underline{88.07} & 81.24 & 83.38 & 86.15 \\
        PlainMamba \cite{PlainMamba} & 55.51 & 51.53 & 52.67 & 58.66 & 55.51 & 64.68 & 52.43 & 53.17 & 43.72 & 65.48 & 52.43 & 47.01 & 82.74 & 79.25 & 84.04 & 73.15 & 76.42 & 84.09 \\
        SCSegamba \cite{SCSegamba} & 64.42 & 63.52 & 62.42 & 66.55 & 64.42 & 60.31 & 82.56 & 81.32 & 80.67 & 84.55 & 82.56 & 83.08 & 69.35 & 69.41 & 79.55 & 64.39 & 71.17 & 86.61 \\
        Fluxamba & \textbf{82.98} & \textbf{82.29} & \textbf{86.27} & \textbf{87.41} & \textbf{82.72} & \textbf{82.89} & \textbf{87.29} & \textbf{86.76} & \textbf{84.87} & \textbf{90.47} & \textbf{86.91} & \textbf{86.85} & \textbf{89.12} & \textbf{89.58} & \textbf{88.53} & \textbf{89.42} & \textbf{89.22} & \textbf{89.87} \\
        \hline 
    \end{tabular}
    \label{table1}
\end{table*}

\begin{table}[htb]
    \caption{Complexity and efficiency analysis. All metrics are evaluated on a $512 \times 512$ input using a single NVIDIA RTX 3090 GPU. Best results are in bold, and second-best results are underlined.}
    \centering
    \setlength\tabcolsep{1.5mm}
    \fontsize{9pt}{11pt}\selectfont
    \begin{tabular}{l|c|c|c|c}
        \hline
        Methods & $\text{FLOPs} \downarrow$ & $\text{Params} \downarrow$ & $\text{Size} \downarrow$ & $\text{FPS} \uparrow$ \\
        \hline
        RIND \cite{RINDNet} & 695.77G & 59.39M & 453MB & \underline{28.46} \\
        CT-crackseg \cite{CT-CrackSeg} & 39.47G & 22.88M & 174MB & 5.94 \\
        SimCrack \cite{SimCrack} & 286.62G & 29.58M & 225MB & \textbf{40.81} \\
        Swin-UMamba \cite{Swin-UMamba} & 616.63G & 23.92M & 91.25MB & 20.20 \\
        VMamba \cite{VMamba} & 395.24G & 19.50M & 65.54MB & 23.06 \\
        VM-UNet \cite{VM-UNet} & 29.42G & 27.25M & 105.92MB & 21.83 \\ 
        PlainMamba \cite{PlainMamba} & 73.36G & 16.72M & 96MB & 22.67 \\
        SCSegamba \cite{SCSegamba} & \underline{18.16G} & \textbf{2.80M} & \underline{37MB} & 9.48 \\
        Fluxamba (Ours) & \textbf{6.25G} & \underline{3.39M} & \textbf{12.92MB} & 24.12 \\
        \hline
    \end{tabular}
    \label{table2}
\end{table}

\subsection{Implementation Details}

\paragraph{Experimental Settings.}
Our framework was implemented using PyTorch 2.4.0 on a workstation equipped with NVIDIA GeForce RTX 3090 GPUs. Input images were resized to a uniform resolution of $512 \times 512$ pixels. To ensure reproducibility, the random seed was fixed at 42.
To improve generalization capabilities on limited geological samples, we applied standard data augmentation techniques, including random vertical/horizontal flips and rotations, during training.
Network parameters were optimized using the AdamW optimizer with an initial learning rate of $1 \times 10^{-5}$ and a weight decay of $0.01$. We employed a polynomial learning rate scheduler (PolyLR) with a power of 0.9 to dynamically decay the learning rate over 200 epochs. To mitigate overfitting, the model checkpoint yielding the highest F1-score on the validation set was rigorously selected for final testing. All models were trained with a batch size of 16 to ensure stable convergence.To rigorously benchmark the inference speed reported in Table II, we measured latency using automatic mixed precision (FP16) with synchronized CUDA execution on a single RTX 3090 GPU. We averaged the results over 1,000 runs following a warm-up period to explicitly exclude data loading and compilation overhead.
\vspace{-10pt}
\paragraph{Comparison Baselines.}
To validate the topological modeling superiority of Fluxamba, we conducted a comprehensive comparative analysis against eight representative State-of-the-Art (SOTA) methods spanning two distinct paradigms. First, to benchmark against established inductive biases, we selected three CNN/Hybrid-based methods: RIND \cite{RINDNet}, CT-crackseg \cite{CT-CrackSeg}, and SimCrack \cite{SimCrack}. Second, to directly assess the advantage of our topology-aware design over rigid scanning mechanisms, we compared against five leading Vision State Space Models (SSMs), including Swin-UMamba \cite{Swin-UMamba}, VMamba \cite{VMamba}, VM-UNet \cite{VM-UNet}, PlainMamba \cite{PlainMamba}, and SCSegamba. This diverse selection ensures a rigorous evaluation of Fluxamba's efficiency against CNNs and its structural preservation capabilities against standard SSMs.

\begin{table*}[tb]
    \caption{Ablation study of components within the SF block. All performance metrics are in percentage (\%). Best results are in bold, and second-best results are underlined.}
    \centering
    \setlength\tabcolsep{2mm} 
    \fontsize{9pt}{11pt}\selectfont 
    \begin{tabular}{cccc|cccccc|ccc} 
        \hline
        PMF & ASG & HSR & HFFU & ODS & OIS & P & R & F1 & mIoU & Params $\downarrow$ & FLOPs $\downarrow$ & Size $\downarrow$ \\
        \hline
        $\times$ & $\times$ & $\times$ & $\times$ & 76.49 & 77.03 & 70.50 & 78.25 & 74.17 & 79.33 & 1.20M & 5.05G & 4.58MB \\
        
        $\checkmark$ & $\times$ & $\times$ & $\times$ & 86.65 & 87.19 & 87.03 & 86.34 & 86.68 & 88.10 & 1.92M & 5.08G & 7.33MB \\
        
        $\times$ & $\checkmark$ & $\times$ & $\times$ & 81.28 & 81.69 & 72.84 & 80.01 & 76.25 & 80.66 & 1.86M & 5.57G & 7.10MB \\
        
        $\times$ & $\times$ & $\checkmark$ & $\times$ & 82.84 & 83.82 & 81.16 & 84.44 & 82.77 & 85.15 & 1.64M & 5.44G & 6.24MB \\

        $\times$ & $\times$ & $\times$ & $\checkmark$ & 81.28 & 81.65 & 79.39 & 83.23 & 81.26 & 84.07 & 1.57M & 5.32G & 5.97MB \\
        
        $\checkmark$ & $\checkmark$ & $\times$ & $\times$ & 87.56 & 88.06 & 87.35 & 87.81 & 87.58 & 88.80 & 2.58M & 5.60G & 9.86MB \\
        
        $\checkmark$ & $\times$ & $\checkmark$ & $\times$ & 87.54 & 88.19 & 86.49 & \underline{88.51} & 87.49 & 88.73 & 2.36M & 5.47G & 8.99MB \\

        $\checkmark$ & $\times$ & $\times$ & $\checkmark$ & \underline{88.23} & \underline{88.71} & \underline{88.16} & 88.31 & \underline{88.23} & \underline{89.32} & 2.29M & 5.34G & 8.73MB \\

        $\times$ & $\checkmark$ & $\checkmark$ & $\times$ & 85.14 & 85.79 & 82.56 & 87.60 & 85.01 & 86.81 & 2.30M & 5.96G & 8.77MB \\
        
        $\times$ & $\checkmark$ & $\times$ & $\checkmark$ & 79.02 & 79.66 & 76.66 & 81.22 & 78.87 & 82.41 & 2.23M & 5.84G & 8.50MB \\

        $\times$ & $\times$ & $\checkmark$ & $\checkmark$ & 85.70 & 86.48 & 83.13 & 88.24 & 85.61 & 87.27 & 2.00M & 5.71G & 7.64MB \\

        $\checkmark$ & $\checkmark$ & $\checkmark$ & $\times$ & 85.75 & 86.52 & 83.67 & 87.84 & 85.71 & 87.34 & 3.02M & 5.99G & 11.52MB \\

        $\checkmark$ & $\checkmark$ & $\times$ & $\checkmark$ & 87.07 & 87.41 & 87.53 & 86.67 & 87.11 & 88.42 & 2.95M & 5.87G & 11.25MB \\

        $\checkmark$ & $\times$ & $\checkmark$ & $\checkmark$ & 87.76 & 88.45 & 86.99 & 88.43 & 87.71 & 88.90 & 2.72M & 5.73G & 10.39MB \\

        $\times$ & $\checkmark$ & $\checkmark$ & $\checkmark$ & 85.24 & 85.99 & 83.53 & 86.82 & 85.15 & 86.92 & 2.66M & 6.23G & 10.16MB \\
        
        $\checkmark$ & $\checkmark$ & $\checkmark$ & $\checkmark$ & \textbf{89.12} & \textbf{89.58} & \textbf{88.53} & \textbf{89.42} & \textbf{89.22} & \textbf{89.87} & 3.39M & 6.25G & 12.92MB\\
        
        \hline
    \end{tabular}
    \label{table3}
\end{table*}
\vspace{-10pt}
\paragraph{Evaluation Protocols.}
We adopted a holistic evaluation protocol assessing both segmentation fidelity and computational efficiency. Following standard benchmarks for linear structure segmentation, we report six quantitative metrics: Precision (P), Recall (R), F1-Score, Optimal Dataset Scale (ODS), Optimal Image Scale (OIS), and mean Intersection over Union (mIoU).
Specifically, ODS and OIS are critical for evaluating boundary continuity under varying probability thresholds $t \in (0, 1)$. ODS represents the best F1-score obtained using a fixed global threshold optimal for the entire dataset:
\begin{equation}
\text{ODS} = \max_{t} \left( \frac{2 \cdot P_t \cdot R_t}{P_t + R_t} \right)
\end{equation}
Conversely, OIS aggregates the F1-scores based on the optimal threshold derived for each individual image $i$:
\begin{equation}
\text{OIS} = \frac{1}{N} \sum_{i=1}^{N} \max_{t} \left( \frac{2 \cdot P_{t,i} \cdot R_{t,i}}{P_{t,i} + R_{t,i}} \right)
\end{equation}
Furthermore, mIoU is calculated as the macro-average of the Intersection over Union for the foreground and background classes:
\begin{equation}
\text{mIoU} = \frac{1}{2} \left( \frac{\text{TP}}{\text{TP} + \text{FP} + \text{FN}} + \frac{\text{TN}}{\text{TN} + \text{FN} + \text{FP}} \right)
\end{equation}
Regarding computational efficiency, we report the number of floating-point operations (FLOPs), learnable parameters (Params), and physical model size (MB) to demonstrate the lightweight nature of our architecture.

\subsection{Comparison with SOTA Methods}

As quantitatively substantiated in Table \ref{table1}, the proposed Fluxamba demonstrates definitive superiority over eight state-of-the-art methods, establishing new performance benchmarks across three geologically diverse datasets. By consistently securing the top rank across all quantitative metrics, our framework evidences the robustness of the Structural Flux Block (SFB) in orchestrating a holistic topology-aware segmentation process.

On planetary surfaces characterized by extremely low signal-to-noise ratio (e.g., LROC-Lineament and LineaMapper), Fluxamba exhibits remarkable capabilities in extracting faint, anisotropic structures. Specifically, on the LROC-Lineament dataset, our model achieves an F1-score of 89.22\%, surpassing the runner-up by a substantial margin of 2.84\%. We attribute this significant gain to the seamless coupling of the Anisotropic Structural Gate (ASG) and the High-Fidelity Focus Unit (HFFU). While the ASG explicitly encodes geometric priors to "lock onto" weak linear traces amidst impact cratering, the HFFU concurrently maximizes the signal-to-noise ratio via dual-polarized modulation. It is this guidance-and-filtering synergy that allows Fluxamba to reconstruct continuous lunar rilles that are otherwise lost by standard CNNs due to background clutter.

In the context of terrestrial geological mapping (GeoCrack), where models face intense textural interference and intricate fracture networks, Fluxamba achieves an exceptional Recall of 87.41\%, significantly outperforming RIND (80.04\%) and CT-crackseg (80.41\%). This superiority validates the collaborative efficacy of the Prior-Modulated Flow (PMF) and the Hierarchical Spatial Regulator (HSR). The PMF effectively rectifies rigid scanning paths to traverse complex rock surfaces, capturing long-range dependencies essential for topological continuity. Complementarily, the HSR aligns these rectified features across scales, ensuring that the fine-grained boundaries of micro-fractures are not eroded during feature aggregation. This flow-and-alignment mechanism ensures that Fluxamba excels in disentangling target fractures from the chaotic lithological background, a feat that rigid SSMs struggle to achieve.

\subsection{Complexity and Efficiency Analysis}

Beyond segmentation fidelity, the feasibility of deployment on resource-limited exploration platforms (e.g., orbital satellites or planetary rovers) depends critically on the model's computational footprint and inference latency. As systematically benchmarked in Table \ref{table2}, we assess the efficiency of Fluxamba against peers by standardizing all inputs to a $512 \times 512$ resolution on a single NVIDIA RTX 3090 GPU.

Fluxamba establishes a new Pareto frontier between model complexity and runtime performance. In terms of theoretical computational cost, our model requires a mere 6.25G FLOPs, representing a reduction of two orders of magnitude compared to heavy-weight Transformers like Swin-UMamba (616.63G FLOPs). Even when compared to the efficient SSM-based competitor SCSegamba, Fluxamba achieves a remarkable 65.6\% reduction in FLOPs (6.25G vs. 18.16G). This substantial decrease in floating-point operations directly translates to lower energy consumption, a primary constraint for onboard hardware.

Crucially, regarding practical inference throughput, Fluxamba achieves a processing speed of 24.12 FPS, exceeding the standard video-rate real-time threshold (24 FPS). This corresponds to an approximate $2.5\times$ acceleration compared to SCSegamba (9.48 FPS), demonstrating that our macro-level rectification strategy avoids the latency bottlenecks inherent in other serialized scanning approaches. While SimCrack exhibits a higher raw FPS (40.81), it incurs an excessive computational overhead (286.62G FLOPs) and a large storage footprint (225MB), rendering it less viable for energy-critical edge devices.

Regarding storage requirements, Fluxamba maintains a lightweight profile tailored for efficient uplink/downlink. Although our parameter count (3.39M) is marginally higher than that of SCSegamba (2.80M), our model benefits from optimized inference weight serialization, resulting in a physical footprint of only 12.92MB. This contrasts significantly with the 37MB reported for SCSegamba \cite{SCSegamba} highlighting the effectiveness of our deployment-oriented optimization. This compactness, combined with real-time inference capability and minimal FLOPs, demonstrates superior feasibility for in-orbit execution where bandwidth, energy, and latency are all critical constraints.

\subsection{Ablation Studies}
\label{sec:ablation}

To rigorously validate the architectural integrity and dissect the specific contributions of each component within Fluxamba, we conducted comprehensive ablation studies on the validation set.
\vspace{-10pt}
\paragraph{Synergistic Efficacy of SFB Components.}
Table \ref{table3} systematically isolates the contribution of each module within the Structural Flux Block (SFB). The baseline model, utilizing a standard Mamba block, yields a limited mIoU of 79.33\%, primarily due to its inability to maintain topological continuity. The integration of the Prior-Modulated Flow (PMF) serves as the architectural foundation, providing the necessary structural connectivity and boosting the mIoU by a substantial 8.77\% to reach 88.10\%. However, relying solely on PMF tends to introduce aliasing artifacts and lacks fine-grained boundary precision. This limitation is effectively addressed by the collaborative integration of the Hierarchical Spatial Regulator (HSR) and the Anisotropic Structural Gate (ASG). As evidenced by comparing the PMF-only variant with the integrated counterpart (PMF+ASG+HSR), the inclusion of these modules elevates performance by explicitly aligning multi-scale features and injecting geometric priors. While their parameter overhead is marginal, their role is indispensable in transforming "coarse connections" into "precise delineations," preventing the erosion of fine-grained fracture details. Finally, the High-Fidelity Focus Unit (HFFU) acts as the ultimate gatekeeper against noise. Removing HFFU from the full model leads to a marked 2.53\% decline in mIoU, confirming that even with perfect topology, noise suppression is non-negotiable for high-fidelity segmentation in signal-starved environments. Consequently, the full Fluxamba architecture achieves the optimal mIoU of 89.87\%, demonstrating that SOTA performance is the result of the synergistic interplay between rectification (PMF), alignment (HSR), and purification (HFFU), achieved with highly efficient parameter utilization (3.39M).

\begin{table}[htb] 
    \caption{Ablation studies with different five-route scanning strategies in the SF block. All values are in percentage (\%). Best results are in bold.}
    \centering
    \setlength\tabcolsep{1.5mm} 
    \fontsize{9pt}{11pt}\selectfont 
    \begin{tabular}{l|cccccc} 
        \hline
        \makecell[l]{Scanning \\ Strategy} & ODS & OIS & P & R & F1 & mIoU \\
        \hline
        Parallel & 79.93 & 80.31 & 78.23 & 78.66 & 78.44 & 82.12 \\
        Diag & 79.35 & 80.17 & 74.97 & 81.16 & 77.94 & 81.78 \\
        ParaSna & 85.18 & 85.83 & 86.05 & 82.85 & 84.42 & 86.37 \\
        DiagSna & 85.32 & 85.97 & 84.56 & 84.49 & 84.49 & 86.42 \\
        SASS & 86.89 & 86.24 & 84.55 & 86.87 & 85.73 & 87.33 \\
        FS2D (Ours) & \textbf{89.12} & \textbf{89.58} & \textbf{88.53} & \textbf{89.42} & \textbf{89.22} & \textbf{89.87} \\
        \hline
    \end{tabular}
    \label{table4}
\end{table}
\vspace{-10pt}
\paragraph{Superiority of Topology-Aware Scanning.}
Table \ref{table4} quantitatively analyzes the impact of token serialization strategies. Rigid strategies (Parallel, Diagonal) fail to capture curvilinear dependencies, limiting performance to $\sim$82\% mIoU due to their discontinuous receptive fields. 
While continuous approaches like Snake scanning improve feature connectivity, they remain spatially agnostic. Crucially, we compare against the SOTA method SASS \cite{SCSegamba}, which employs a fixed combination of parallel and diagonal snake paths. Although SASS achieves a competitive 87.33\% mIoU by increasing directional coverage, it remains bound by predefined trajectories. 
In contrast, our Four-directional Selective 2D Scan (FS2D) outperforms SASS by 2.54\% in mIoU and 3.49\% in F1-score. 
The decisive advantage lies in the \textit{dynamic rectification} mechanism: rather than relying on a static bank of scanning paths, FS2D utilizes geometric priors from the ASG to adaptively re-weight the information flux. This explicitly aligns feature propagation with the target's intrinsic topology, proving that content-aware gating is superior to static multi-path combinations.

\begin{table}[htb] 
    \caption{Ablation study of different segmentation heads. All values are in percentage (\%). Best results are in bold.}
    \centering
    \setlength\tabcolsep{1.5mm} 
    \fontsize{9pt}{11pt}\selectfont 
    \begin{tabular}{l|cccccc} 
        \hline
        Seg Head & ODS & OIS & P & R & F1 & mIoU \\
        \hline
        UNet      & 88.41 & 88.81 & 86.92 & \textbf{89.52} & 88.23 & 89.29 \\
        Ham       & 87.11 & 87.34 & 86.85 & 89.11 & 87.97 & 89.11 \\
        SegFormer & 86.59 & 87.21 & 85.46 & 86.30 & 85.88 & 87.47 \\
        MFS       & 87.36 & 87.26 & 86.61 & 88.13 & 87.36 & 88.48 \\
        BMF       & \textbf{89.12} & \textbf{89.58} & \textbf{88.53} & 89.42 & \textbf{89.22} & \textbf{89.87} \\
        \hline
    \end{tabular}
    \label{table5}
\end{table}

\vspace{-10pt}
\paragraph{Impact of Segmentation Heads.}
Finally, Table \ref{table5} assesses the effectiveness of the decoding stage. While general-purpose heads like SegFormer and U-Net achieve competitive results, they often struggle with boundary blurring in complex geological scenarios. Our Boundary-Modulated Fusion (BMF) head surpasses the widely used SegFormer head by 2.40\% in mIoU and outperforms the U-Net head by roughly 1\% in F1-score. These results indicate that the BMF's residual injection mechanism—which re-introduces high-resolution spatial details from the shallowest stage—is essential for recovering the pixel-level acuity of linear boundaries, ensuring high-precision delineation with negligible computational overhead.

\begin{table}[htb]
    \caption{DETAILED ARCHITECTURE SPECIFICATIONS. We instantiate four Fluxamba variants by scaling the number of SFB blocks $[d_1, d_2, d_3, d_4]$ across the four encoder stages. "Size" denotes the physical model storage footprint.}
    \label{tab:model_specs}
    \centering
    \setlength\tabcolsep{1mm} 
    \fontsize{9pt}{11pt}\selectfont 
    
    \begin{tabular}{l|c|ccc}
        \hline
        \multirow{2}{*}{Variant} & Depth Config. & \multicolumn{3}{c}{Resource Consumption} \\
        \cline{3-5}
         & $[d_1, d_2, d_3, d_4]$ & Params $\downarrow$ & FLOPs $\downarrow$ & Size $\downarrow$ \\
        \hline
        Fluxamba-T & $[1, 1, 2, 1]$ & \textbf{3.39M} & \textbf{6.23G} & \textbf{12.92MB} \\
        Fluxamba-S & $[2, 2, 3, 2]$ & 11.79M & 31.30G & 44.99MB \\
        Fluxamba-B & $[2, 2, 4, 2]$ & 22.26M & 57.01G & 84.92MB\\
        Fluxamba-L & $[2, 3, 6, 3]$ & 69.53M & 144.11G & 265.22MB \\
        \hline
    \end{tabular}
\end{table}

\begin{table}[htb]
   \caption{SCALABILITY ANALYSIS OF FLUXAMBA VARIANTS ACROSS DIFFERENT DEPTH CONFIGURATIONS. ALL VALUES ARE IN PERCENTAGE (\%). BEST RESULTS ARE IN BOLD.}
    \label{tab:ablation_variants_perf}
    \centering
    \setlength\tabcolsep{1.5mm}
    \fontsize{9pt}{11pt}\selectfont
    
    \begin{tabular}{l|cccccc}
        \hline
        Variant & ODS & OIS & P & R & F1 & mIoU \\
        \hline
        Fluxamba-T & 81.02 & 81.65 & 81.81 & 84.20 & 83.00 & 84.13 \\
        Fluxamba-S & 81.12 & 81.76 & 81.95 & 84.45 & 83.00 & 84.34 \\
        Fluxamba-B & 81.78 & 81.95 & 82.10 & 84.90 & 83.12 & 84.44 \\
        Fluxamba-L & \textbf{81.99} & \textbf{82.00} & \textbf{82.23} & \textbf{85.20} & \textbf{83.32} & \textbf{84.67} \\
        \hline
        Gain ($T \to L$) & +0.97 & +0.35 & +0.42 & +1.00 & +0.32 & +0.54 \\
        \hline
    \end{tabular}
\end{table}

\vspace{-10pt}
\paragraph{Scalability and Pareto Efficiency Analysis.}
To rigorously determine the optimal model capacity for resource-constrained deployment, we investigate the scalability of Fluxamba by systematically varying the network depth. As detailed in Table \ref{tab:model_specs}, we instantiated four variants—Tiny (T), Small (S), Base (B), and Large (L)—by scaling the number of cascaded Structural Flux Blocks (SFB) across the four encoder stages. This configuration allows us to monitor the trajectory of resource consumption, which spans from a lightweight 3.39M parameters to a heavy-weight 69.53M.

Correlating these specifications with the performance metrics in Table \ref{tab:ablation_variants_perf} reveals a stark law of diminishing returns. The transition from Tiny to Large incurs a massive computational penalty: parameter count and FLOPs surge by approximately $20\times$ and $23\times$, respectively. However, this exponential growth in complexity yields only a negligible performance benefit, with mIoU improving by a mere 0.54\% (84.13\% $\rightarrow$ 84.67\%). 

This disparity underscores a critical architectural insight: the topological modeling superiority of Fluxamba stems from the efficient design of the SFB unit itself, rather than brute-force depth stacking. Consequently, the Fluxamba-Tiny variant establishes the optimal Pareto frontier, delivering high-fidelity segmentation with minimal overhead, making it the definitive choice for onboard processing in deep space missions.

\begin{figure*}[ht]
\centering
\includegraphics[width=0.8\textwidth]{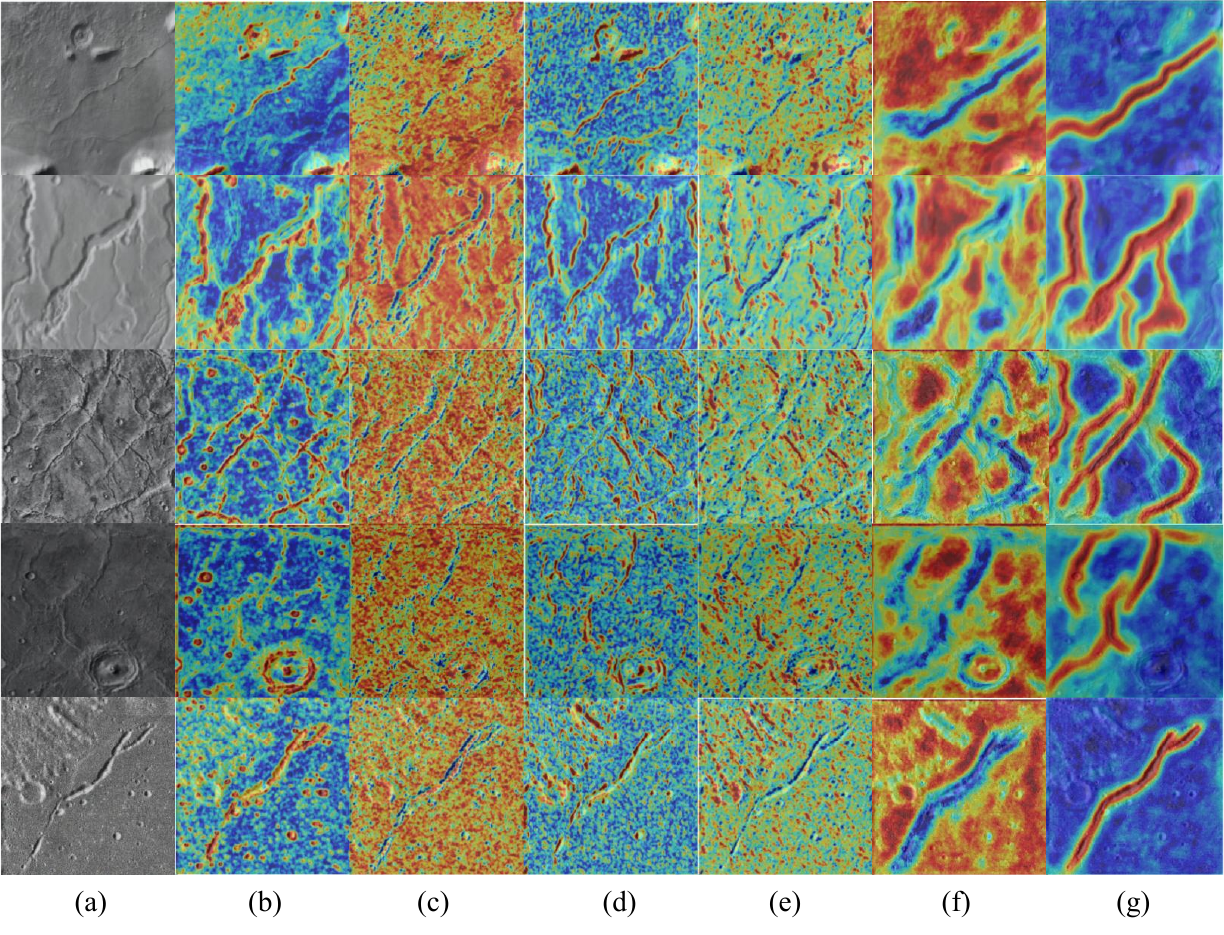} 
\caption{
Visualization of the layer-wise feature evolution mechanism, illustrating the Dynamic Polarity Oscillation phenomenon.
(a) Original Imagery.
(b) ASG Output: The module acts as a geometric primer, generating a coarse positive activation map (red) to localize fracture orientations.
(c) PMF Output: Features spontaneously invert to negative polarity (blue/cyan) to facilitate connectivity establishment amidst the dominant background.
(d) HSR Output: The feature polarity temporarily reverts to positive (red), visually confirming the spatial anchoring process that realigns semantic flows with the physical grid.
(e) HFFU Output: A second inversion occurs where the background variance is homogenized into a high-energy saturated state (red/yellow), effectively isolating targets as pure topological voids (blue).
(f) BMF Output: The DySample operator restores spatial resolution while preserving the sharp gradients of the feature valleys.
(g) Final Prediction: The segmentation head acts as a semantic rectifier to map the manifold back to the positive probability space.
Note the distinct alternating activation states (Red$\rightarrow$Blue$\rightarrow$Red$\rightarrow$Blue$\rightarrow$Red) that progressively enhance the signal-to-noise ratio.
}
\label{figure5}
\end{figure*}

\subsection{Evolution of Learned Representations}
\label{sec:vis_feature}

To investigate the internal representational dynamics of Fluxamba, Fig. \ref{figure5} visualizes the sequential transformation of the feature manifold, strictly following established interpretability protocols \cite{zeiler2014visualizing}. This progression reveals a distinct \textit{Alternating Polarity Phenomenon}, where the network iteratively shifts between positive (spatial-dominant) and negative (semantic-dominant) activation states. We interpret this oscillation not as a random artifact, but as an emergent behavior optimizing the trade-off between geometric fidelity and noise suppression.
\vspace{-10pt}
\paragraph{Geometric Priming and Connectivity Establishment (Fig. \ref{figure5}(b)-(c)).}
The evolution commences with the Anisotropic Structural Gate (ASG) in Fig. \ref{figure5}(b). Functioning as a geometric encoder, the ASG generates a coarse attention map where linear structures are positively activated (red). This effectively establishes the initial spatial localization, anchoring fracture orientations. However, as features propagate through the Prior-Modulated Flow (PMF) (Fig. \ref{figure5}(c)), a marked inversion occurs. We attribute this to the statistical properties of normalization layers (e.g., LayerNorm \cite{ba2016layer}) when processing sparse geological targets: the dominant background features are shifted toward the positive mean, consequently forcing the sparse lineaments into low-energy ``valleys'' (blue). Within this inverted manifold, the PMF exploits the contrast to prioritize long-range connectivity, bridging fragmented segments to form a continuous topological skeleton amidst high-energy background variance.
\vspace{-10pt}
\paragraph{Spatial Anchoring and Manifold Purification (Fig. \ref{figure5}(d)-(e)).}
A critical realignment occurs at the Hierarchical Spatial Regulator (HSR) stage (Fig. \ref{figure5}(d)), where feature polarity reverts to a positive state (red). This observation empirically validates the HSR's role as a multi-scale spatial anchor. Depending on the encoder depth, the HSR corrects geometric distortions introduced by the deep flow: it either reinjects local spatial gradients via the Lightweight Modulation Refinement (LMR) in shallow stages or enforces global semantic coherence via the Global Transformer Reorganizer (GTR) in deep stages. This dual-mechanism ensures that abstract features remain aligned with the physical spatial grid.
Subsequently, to maximize the Signal-to-Noise Ratio (SNR), the High-Fidelity Focus Unit (HFFU) triggers a second inversion (Fig. \ref{figure5}(e)). By acting as a soft spectral gate, the HFFU homogenizes the background into a high-energy saturated state (red/yellow) while pushing targets back into deep ``valleys'' (blue/cyan). This effectively minimizes the entropy of the surrounding lithological texture, isolating the target structure as a topological void.
\vspace{-10pt}
\paragraph{Valley-Preserving Restoration and Semantic Rectification (Fig. \ref{figure5}(f)-(g)).}
The decoding phase preserves this purified, inverted manifold structure. The Boundary-Modulated Fusion (BMF) head (Fig. \ref{figure5}(f)) performs upsampling while retaining the negative polarity (blue). This suggests that the content-aware DySample operator \cite{liu2023learning} is optimized to propagate the sharp gradients of these ``feature valleys,'' avoiding the blurring artifacts often associated with premature rectification. Finally, the segmentation head (Fig. \ref{figure5}(g)) functions as a semantic probabilistic mapper, projecting these deep feature valleys back into the positive probability space to yield the final prediction, thereby completing the topology-aware extraction workflow.

\begin{figure*}[htb]
\centering
\includegraphics[width=1\textwidth]{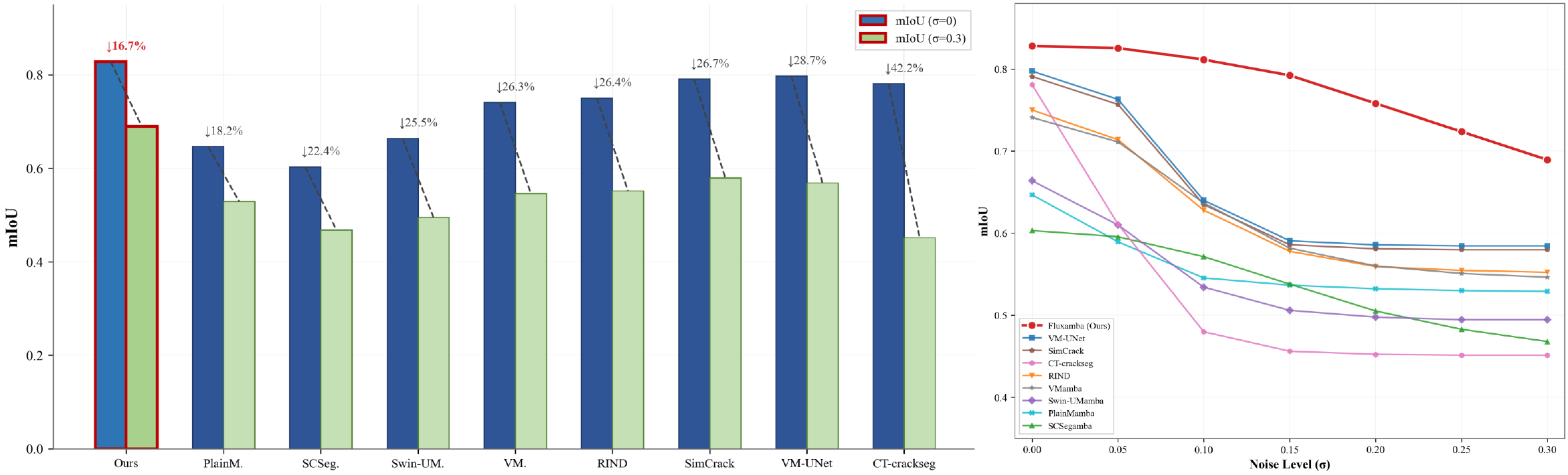} 
\caption{
Quantitative evaluation of algorithmic robustness against stochastic signal degradation.
(a) Performance Drop Rate Analysis: Comparison of relative mIoU decline from clean ($\sigma=0$) to noisy ($\sigma=0.3$) conditions. Fluxamba (red bar) exhibits the highest resilience with only a 16.7\% drop, whereas texture-dependent baselines like CT-crackseg suffer over 40\% degradation.
(b) Stability Trajectories: mIoU evolution under increasing noise intensities. Fluxamba maintains a stable performance envelope, significantly outperforming SOTA methods. This stability ensures the preservation of topological connectivity in degraded sensing environments (e.g., SAR speckle), which is vital for reliable geological hazard assessment.
}
\label{fig:robustness}
\end{figure*}

\subsection{Reliability Analysis for Hazard Assessment}
\label{sec:robustness}

In remote sensing-based geological hazard analysis—such as post-earthquake rupture mapping and landslide precursor monitoring—image quality is inevitably compromised by stochastic perturbations, including atmospheric scattering (e.g., dust, smoke) or sensor-inherent artifacts (e.g., thermal noise, SAR speckle \cite{argenti2013tutorial}). Under such signal-degraded conditions, traditional segmentation models often yield fragmented predictions, severing the topological continuity of critical fault lines. This fragmentation poses a severe systemic risk: in seismic hazard assessment, a discontinuous fault trace can lead to a dangerous underestimation of the potential rupture length and moment magnitude \cite{wells1994new}.

To rigorously evaluate the operational reliability of Fluxamba in these mission-critical scenarios, we conducted stress tests by injecting Gaussian noise of varying intensities ($\sigma \in [0, 0.3]$) into the test samples \cite{hendrycks2019benchmarking}. As visualized in Fig. \ref{fig:robustness}, the results reveal two decisive advantages for rapid disaster response:
\vspace{-10pt}
\paragraph{Resilience to Catastrophic Signal Degradation.}
Fig. \ref{fig:robustness}(a) quantifies the performance retention capability across varying methods. When subjected to severe noise ($\sigma=0.3$), standard CNN and Transformer baselines suffer a catastrophic performance collapse. For instance, CT-crackseg \cite{CT-CrackSeg} exhibits a precipitous drop of 42.2\% in mIoU. This vulnerability indicates an over-reliance on local high-frequency textures, which are easily corrupted by stochastic noise.
In stark contrast, Fluxamba demonstrates exceptional resilience, restricting the performance degradation to a minimal 16.7\%. This represents the lowest drop rate among all comparison methods. This stability implies that our architecture effectively decouples structural semantics from local signal variance, ensuring consistent delineation even when the signal-to-noise ratio approaches critical limits.
\vspace{-10pt}
\paragraph{Topological Integrity via Synergistic Defense.}
The mIoU stability trajectories are plotted in Fig. \ref{fig:robustness}(b). While competing methods (e.g., VM-UNet \cite{VM-UNet}, Swin-UMamba \cite{Swin-UMamba}) show a rapid exponential decay as noise intensity increases, Fluxamba maintains a significantly flatter trajectory, consistently outperforming the nearest competitor by a wide margin across the entire noise spectrum.
This phenomenon validates the mechanistic efficacy of the Structural Flux Block (SFB) as a multi-stage defense system. Specifically, the Anisotropic Structural Gate (ASG) first selectively captures geometric priors, rejecting non-linear noise patterns. Subsequently, the Prior-Modulated Flow (PMF) aggregates global anisotropic context, enabling the model to infer (rather than hallucinate) the latent topology of linear features even when local pixel evidence is ambiguous. Finally, the High-Fidelity Focus Unit (HFFU) functions as a robust spectral filter, attenuating high-frequency noise components \cite{wang2020high} before the BMF head reconstructs the precise boundary. This holistic robustness renders Fluxamba highly suitable for processing degraded imagery, such as speckle-noise-affected SAR data or low-light optical imagery, which are prevalent in time-critical disaster response scenarios.

\begin{figure*}[htb]
\centering
\includegraphics[width=0.82\textwidth]{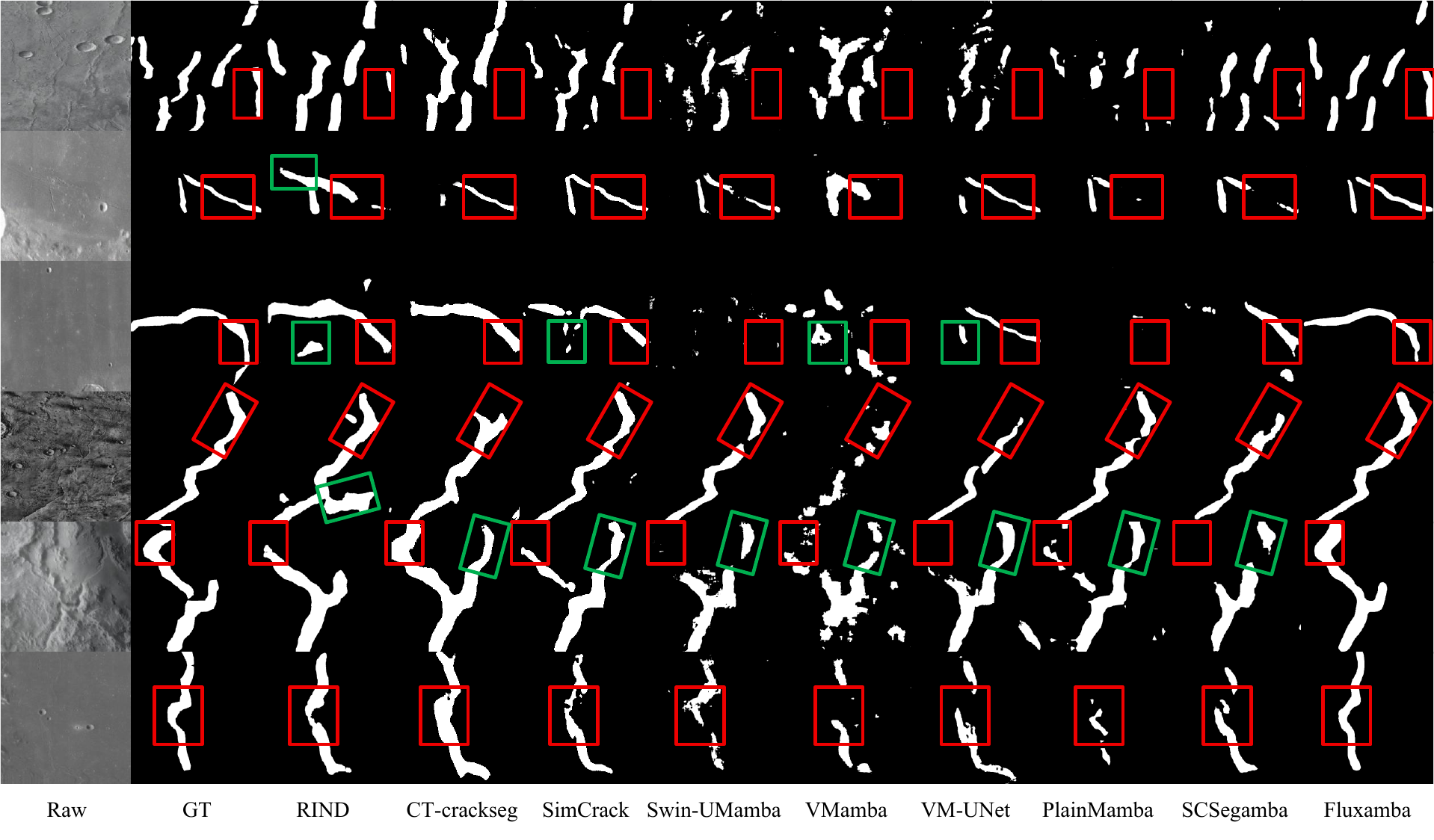}
\caption{
Qualitative segmentation comparison on the LROC-Lineament dataset (Lunar).
Red boxes (Sensitivity to Low-Contrast Targets): Highlight faint, shadowed rilles that typically cause gradient vanishing in baseline methods (e.g., VMamba). Fluxamba successfully recovers these signal-starved structures, demonstrating superior feature extraction capability.
Green boxes (Suppression of Illumination Artifacts): Indicate regions where crater rims and lighting variations induce commission errors in competing models. Fluxamba effectively filters these non-structural distractors.
}
\label{fig:vis_lroc}
\end{figure*}

\begin{figure*}[htb]
\centering
\includegraphics[width=0.82\textwidth]{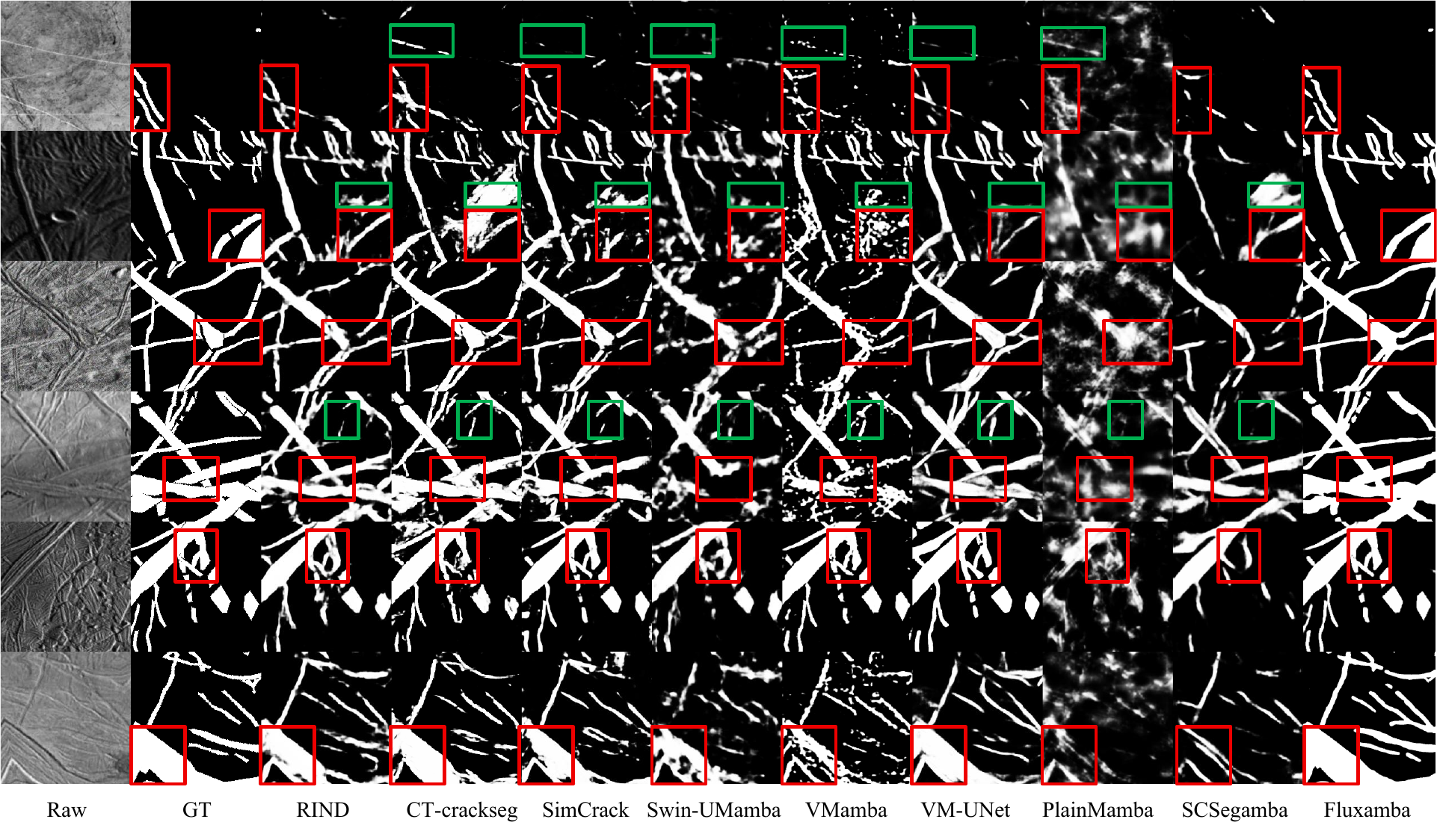}
\caption{
Visual assessment on the LineaMapper dataset (Europa), characterized by intricate ridge networks.
Red boxes (Topological Connectivity): Focus on subtle ridge intersections and cross-cutting relationships. While competitors yield fragmented predictions (broken topology), Fluxamba leverages anisotropic flow to bridge these gaps, preserving global continuity.
Green boxes (Rejection of Textural Artifacts): Mark areas where CNN-based methods (e.g., RIND) hallucinate "blob-like" predictions due to ice shell texture. Fluxamba maintains razor-sharp boundaries, validating its contrastive encoding strategy.
}
\label{fig:vis_europa}
\end{figure*}

\begin{figure*}[htb]
\centering
\includegraphics[width=0.82\textwidth]{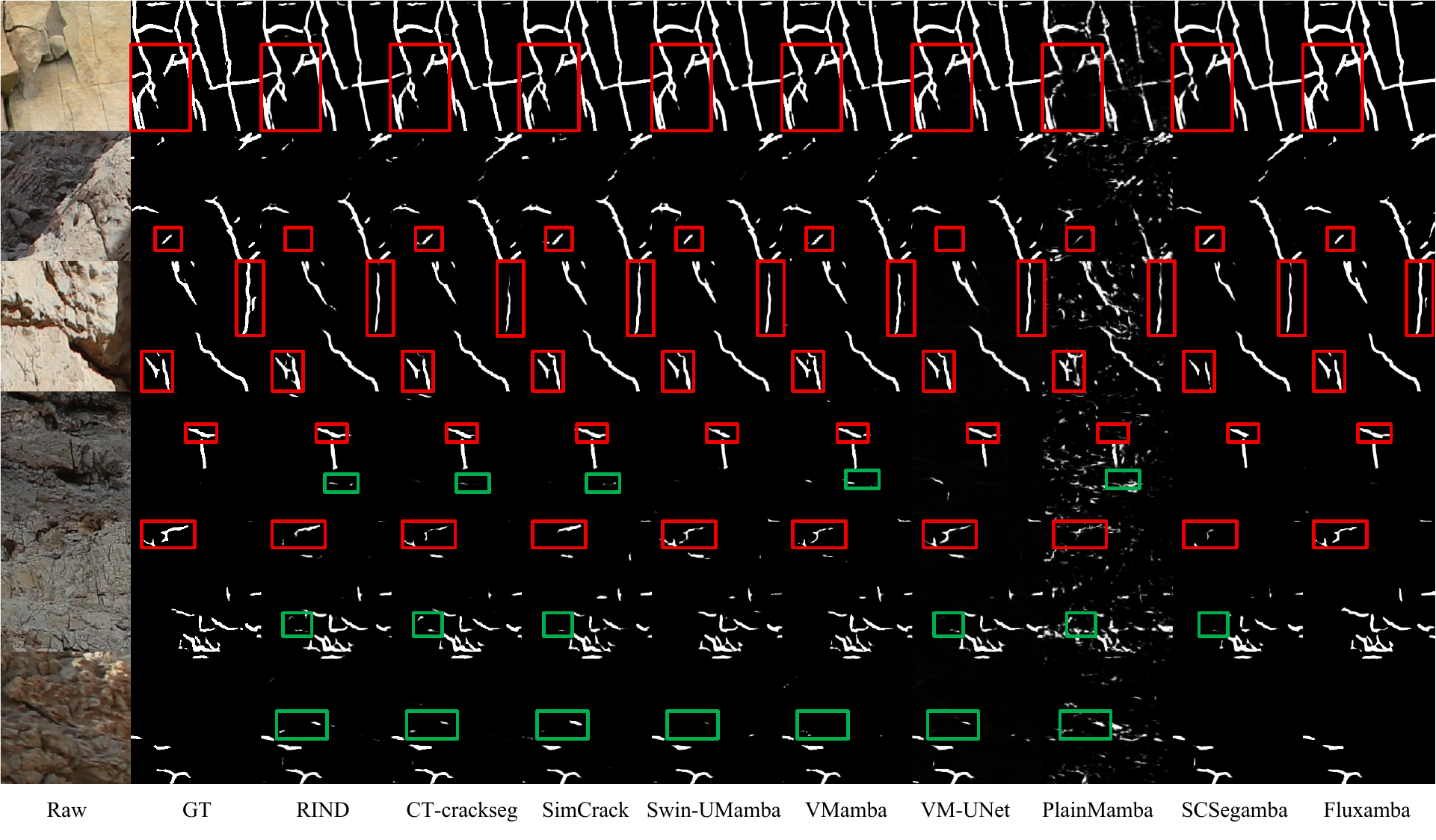}
\caption{
Comparative visualization on the GeoCrack dataset (Terrestrial), testing robustness against intense heterogeneity.
Red boxes (Micro-Structure Delineation): Highlight fine-grained fracture tips and micro-cracks. Fluxamba accurately parses these high-frequency details that are often eroded by pooling operations in other methods.
Green boxes (Robustness to Lithological Noise): Highlight regions where natural rock grain mimics fracture morphology. Fluxamba exhibits superior specificity, effectively distinguishing structural lineaments from stochastic lithological clutter.
}
\label{fig:vis_geocrack}
\end{figure*}

\subsection{Qualitative Analysis}
\label{sec:qualitative}

To provide a perceptual assessment of segmentation fidelity, we visualize representative results in Fig. \ref{fig:vis_lroc}, \ref{fig:vis_europa}, and \ref{fig:vis_geocrack}. These qualitative comparisons explicitly highlight the trade-off between sensitivity (recall) and specificity (precision), revealing how Fluxamba effectively overcomes the prevalent pitfalls of omission errors (missed detections due to weak signals) and commission errors (false alarms due to noise) that plague current SOTA methods.
\vspace{-10pt}
\paragraph{Sensitivity in Signal-Starved Regimes (Fig. \ref{fig:vis_lroc}).}
The LROC-Lineament dataset \cite{Lineament} serves as a stress test for detecting faint structures under extreme illumination variations. As indicated by the red boxes in Fig. \ref{fig:vis_lroc}, baseline models (e.g., Swin-UMamba \cite{Swin-UMamba}) consistently fail to resolve subtle lunar rilles obscured by shadows, resulting in severe topological disconnection. This failure stems from their reliance on strong local gradients, which are absent in such signal-starved regions. In contrast, Fluxamba demonstrates exceptional sensitivity, successfully recovering these low-contrast targets. While the overall topology-aware framework ensures feature continuity, this specific capability is critically enabled by the High-Fidelity Focus Unit (HFFU). By functioning as a spectral filter, the HFFU amplifies weak structural harmonics against the dominant regolith noise. Conversely, the green boxes reveal that Fluxamba effectively suppresses illumination artifacts (e.g., crater rims) that competitors mistake for fractures, verifying the robustness of our noise-filtering mechanism.
\vspace{-10pt}
\paragraph{Topological Completeness vs. Fragmentation (Fig. \ref{fig:vis_europa}).}
On the LineaMapper dataset \cite{Lineamapper} (Fig. \ref{fig:vis_europa}), the primary challenge lies in resolving the intricate connectivity of cryovolcanic networks. Competing methods often struggle to capture global dependencies, leading to fragmented predictions where subtle ridge connections are broken (red boxes). Fluxamba outperforms these baselines by maintaining a coherent global context. Crucially, the Prior-Modulated Flow (PMF) plays a pivotal role in bridging these contextual gaps. By dynamically rectifying the information flux along the intrinsic geometry of the ridges, the PMF ensures global connectivity even across complex intersections. Furthermore, while baselines like RIND \cite{RINDNet} generate "blob-like" artifacts in texture-rich regions (green boxes), Fluxamba produces clean, razor-sharp boundaries, demonstrating that our contrastive encoding strategy effectively distinguishes true topological features from background ice-shell clutter.
\vspace{-10pt}
\paragraph{Specificity amidst Lithological Interference (Fig. \ref{fig:vis_geocrack}).}
The GeoCrack dataset \cite{Geocrack} (Fig. \ref{fig:vis_geocrack}) tests the model's discriminative power against intense lithological heterogeneity. A prevalent failure mode among CNN-based methods (e.g., CT-crackseg \cite{CT-CrackSeg}) is the misclassification of natural rock grain or mineral veins as fractures, resulting in massive commission errors (green boxes). Fluxamba exhibits superior specificity, filtering out these textural distractions while accurately delineating micro-fractures (red boxes). This precision is largely attributed to the Anisotropic Structural Gate (ASG). By utilizing geometric priors to enforce strict structural constraints, the ASG acts as a shape-sensitive filter, rejecting non-linear stochastic textures that lack topological significance, thereby allowing the network to focus exclusively on structural lineaments.

\section{Discussion}

\subsection{Training Dynamics of Non-Residual Filtering}
\label{sec:convergence}

A theoretical concern regarding the High-Fidelity Focus Unit (HFFU) is the intentional omission of residual connections ($\mathbf{y} = \mathbf{x} + \mathcal{F}(\mathbf{x})$). While residual learning is traditionally standard for preventing gradient vanishing \cite{he2016deep}, we posit that for signal-starved planetary segmentation, breaking the identity mapping is a requisite trade-off to physically block the propagation of high-frequency noise.

\begin{figure}[t]
\centering
\includegraphics[width=0.9\columnwidth]{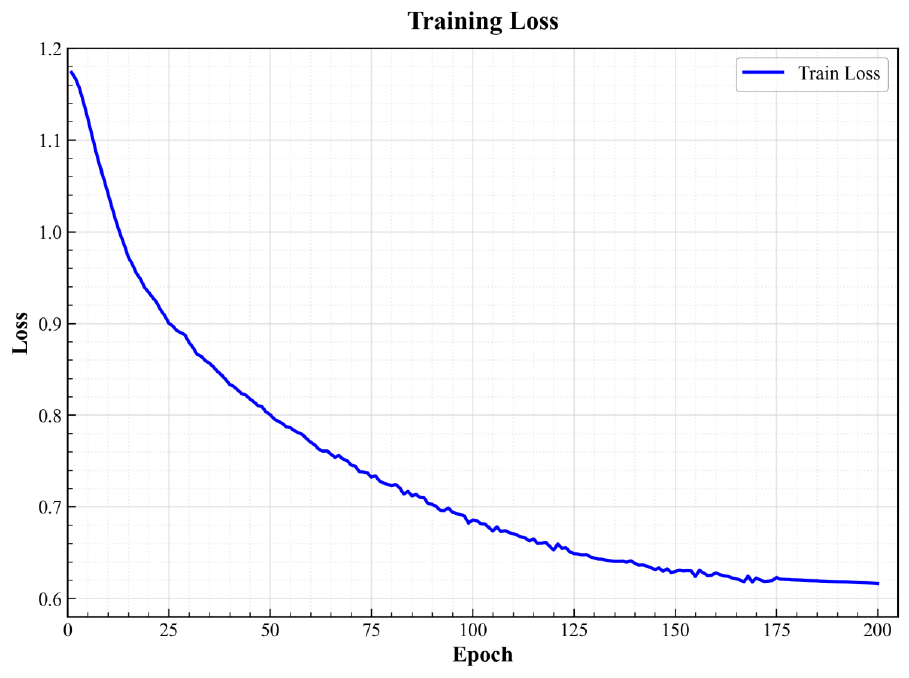}
\caption{
Training loss trajectory of Fluxamba over 200 epochs. Despite minimizing residual connections in the HFFU to enforce rigorous noise filtering, the model (Blue Curve) exhibits a smooth, monotonic convergence profile. The rapid initial descent and subsequent stable plateau empirically verify that the architecture maintains healthy gradient flow, effectively dispelling concerns regarding optimization instability.
}
\label{fig:loss_curve}
\end{figure}

To empirically validate the optimization stability of this "pure filtering" design, we analyze the training loss trajectory in Fig. \ref{fig:loss_curve}.

As evidenced by the blue curve in Fig. \ref{fig:loss_curve}, Fluxamba exhibits a smooth, monotonic decay devoid of the oscillatory instability often observed in ill-conditioned networks. The loss drops rapidly in the initial 50 epochs, indicating effective gradient backpropagation, and progressively stabilizes at a low asymptotic value. This confirms that the structural priors provided by the preceding SFB components establish a sufficiently robust feature manifold, allowing the HFFU to function as a "hard" spectral gate without hindering global trainability. Thus, our design successfully trades the "safety" of residual shortcuts for the "purity" of noise suppression, a strategy validated by the robust convergence behavior.

\subsection{Generalization via Geometric-Radiometric Decoupling}
To validate intrinsic robustness, we conducted a rigorous zero-shot transfer experiment, utilizing the model trained solely on the LROC-Lineament (Lunar) \cite{Lineament} dataset to perform direct inference on the LineaMapper (Europa) \cite{Lineamapper} dataset without fine-tuning.

\begin{table}[htb]
    \caption{Quantitative comparison of Zero-Shot Transfer performance. Models trained on LROC-Lineament were directly evaluated on the unseen LineaMapper dataset. All values are in percentage (\%). Best results are in bold, and second-best results are underlined.}
    \centering
    \setlength\tabcolsep{1.2mm}
    \fontsize{9pt}{11pt}\selectfont
    \begin{tabular}{l|cccccc}
        \hline
        Model & ODS & OIS & P & R & F1 & mIoU \\
        \hline
        RIND \cite{RINDNet} & 43.69 & 45.65 & 40.93 & 13.47 & 20.27 & 39.71 \\
        CT-crackseg \cite{CT-CrackSeg} & \underline{46.48} & \underline{51.32} & 43.25 & 48.14 & 45.57 & 47.60 \\
        SimCrack \cite{SimCrack} & 44.68 & 48.51 & 50.17 & 35.31 & 41.45 & 47.26 \\
        Swin-UMamba \cite{Swin-UMamba} & 43.71 & 47.81 & \underline{52.88} & 28.39 & 37.02 & 46.63 \\
        VMamba \cite{VMamba} & 44.30 & 46.76 & 50.87 & 22.45 & 31.15 & 44.63 \\
        VM-UNet \cite{VM-UNet} & 44.90 & 50.77 & 43.43 & \underline{49.59} & \underline{46.31} & \underline{48.18} \\
        PlainMamba \cite{PlainMamba} & 43.71 & 43.00 & 48.83 & 6.51 & 11.89 & 40.21 \\
        SCSegamba \cite{SCSegamba} & 43.71 & 46.32 & 50.76 & 22.10 & 30.80 & 44.40 \\
        Fluxamba (Ours) & \textbf{51.40} & \textbf{54.66} & \textbf{53.16} & \textbf{49.81} & \textbf{51.30} & \textbf{51.12} \\
        \hline
    \end{tabular}
    \label{tab:zero_shot}
\end{table}
\vspace{-10pt}
\paragraph{Empirical Transferability.}
Despite the substantial domain shift in surface radiometry—from regolith-covered lunar terrain to high-albedo Europan ice shells—Fluxamba demonstrates superior transferability. As presented in Table \ref{tab:zero_shot}, even without seeing a single Europan sample, our model outperforms the strongest baseline (VM-UNet) by margins of +2.94\% in mIoU and +4.99\% in F1-score. This indicates that the learned representations are not strictly bound to the source domain's texture statistics.
\vspace{-10pt}
\paragraph{Theoretical Insight.}
This generalization capability suggests that the Structural Flux Block (SFB) achieves effective geometric-radiometric decoupling. By utilizing the Anisotropic Structural Gate (ASG) as a shape-sensitive filter, the network suppresses domain-specific high-frequency textures (e.g., albedo variations) while selectively amplifying domain-invariant structural topologies. This implies that Fluxamba captures the intrinsic morphology of fractures rather than overfitting to surface appearance, a trait pivotal for planetary missions where training data for new target bodies is unavailable a priori.

\subsection{Feasibility for On-board Deployment}
Deep space exploration imposes stringent constraints on Size, Weight, and Power (SWaP). Our architecture addresses these by optimizing both computational complexity and practical inference throughput.
\vspace{-10pt}
\paragraph{Redefining the Efficiency-Accuracy Paradigm.}
As presented in Table \ref{table2}, Fluxamba establishes a new Pareto frontier. Compared to heavy-weight Transformers like Swin-UMamba (616G FLOPs), it reduces the computational burden by two orders of magnitude (6.25G FLOPs) while achieving higher accuracy.
Crucially, this theoretical efficiency translates into tangible runtime performance: Fluxamba achieves a processing speed of 24.12 FPS, satisfying real-time requirements. This confirms that the macro-level rectification strategy effectively circumvents the quadratic bottlenecks of Attention mechanisms without incurring the latency penalties of complex multi-branch CNNs, making high-throughput execution feasible on legacy space-grade processors \cite{antunes2025fpga}.
\vspace{-10pt}
\paragraph{Maximizing Scientific Yield.}
This high-fidelity, low-latency characteristic directly addresses the bottleneck of limited downlink bandwidth. By enabling in-situ data reduction—converting raw gigapixel imagery into vectorized structural maps onboard—Fluxamba allows spacecraft to transmit only scientifically high-value topological data rather than bulky raster images \cite{rabideau2025planning}, thereby maximizing the scientific yield per downlink bit. Furthermore, the confirmed real-time capability ($>24$ FPS) enables next-generation autonomy in latency-critical scenarios, such as hazard avoidance during descent, where millisecond-level decisions are vital for mission safety.

\subsection{Limitations and Future Prospects}
Despite establishing a new state-of-the-art, certain limitations warrant investigation. First, while the PMF provides rotational robustness via vector synthesis, discretization errors may still arise when modeling fractures with extremely high-frequency curvature changes (e.g., sharp zig-zags) exceeding the resolution of the four cardinal basis vectors. Second, the current implementation is limited to 2D projections, whereas fracture networks are inherently 3D structures.

Future work will focus on: (1) Extending the topology-aware flux mechanism to 3D volumetric seismic interpretation \cite{wu2019faultseg3d}; and (2) Investigating self-supervised pre-training (e.g., Masked Image Modeling) to mitigate annotation scarcity in specialized remote sensing domains \cite{cong2022satmae}.

\section{Conclusion}

In this article, we presented Fluxamba, a topology-aware architecture designed to address the fundamental mismatch between the rigid scanning mechanisms of standard State Space Models and the curvilinear complexity of geological features. 
Instead of relying on fixed, axis-aligned trajectories, our proposed Structural Flux Block (SFB) introduces a dynamic rectification paradigm. This mechanism explicitly aligns the information flow with the intrinsic geometry of the targets, effectively mitigating the feature fragmentation observed in conventional SSMs while preserving continuity in signal-starved environments.

Extensive experiments on three diverse benchmarks (LROC-Lineament, LineaMapper, and GeoCrack) demonstrate that Fluxamba establishes a new state-of-the-art. 
Most notably, it redefines the efficiency-accuracy trade-off essential for planetary exploration: by achieving a real-time inference speed of over 24 FPS with only 3.39M parameters, our model delivers segmentation fidelity comparable to heavy-weight Transformers but with significantly reduced computational overhead. 
This work validates the feasibility of high-performance onboard processing, laying a solid foundation for next-generation autonomous hazard assessment systems. Future research will focus on extending this topology-aware mechanism to 3D volumetric domains and exploring self-supervised learning to mitigate annotation scarcity.

{\small
\bibliographystyle{ieee_fullname}
\balance
\bibliography{refs}
}

\end{document}